\documentclass[10pt,twocolumn,letterpaper]{article}

\usepackage{wacv}              

\usepackage{graphicx}
\usepackage{amsmath}
\usepackage{amssymb}
\usepackage{booktabs}
\usepackage{amsfonts} 
\usepackage{amssymb}
\usepackage{tabularx}
\usepackage{wrapfig}
\usepackage{caption}
\usepackage{array}
\usepackage{amsthm}
\usepackage{amssymb}
\usepackage{graphics}
\usepackage{graphicx}
\usepackage{enumitem}
\usepackage{multirow}
\usepackage{wasysym}
\usepackage{amssymb}
\usepackage{soul}
\usepackage{svg}
\usepackage{tikz}
\usepackage{wasysym}
\usepackage{xspace}
\usepackage{siunitx}
\usepackage{algorithm}
\usepackage{algorithmicx}
\usepackage[noend]{algpseudocode}
\usepackage{xparse}
\usepackage{dsfont}
\usepackage{mathtools}
\usepackage{colortbl}
\usepackage{float}
\usepackage[accsupp]{axessibility}

\let\olduparrow\uparrow
\renewcommand{\uparrow}[1][1pt]{%
  \mathrel{\raisebox{#1}{$\olduparrow$}}%
}

\newcommand{\ours}{DPA\xspace}

\usepackage[pagebackref,breaklinks,colorlinks]{hyperref}

\usepackage[capitalize]{cleveref}
\crefname{section}{Sec.}{Secs.}
\Crefname{section}{Section}{Sections}
\Crefname{table}{Table}{Tables}
\crefname{table}{Tab.}{Tabs.}

\def\wacvPaperID{2287} 
\def\confName{WACV}
\def\confYear{2025}

\begin{document}

\title{\ours: Dual Prototypes Alignment for Unsupervised Adaptation of Vision-Language Models}

\author{
Eman Ali$^{1,2}$
\qquad
Sathira Silva$^{1}$
\qquad
Muhammad Haris Khan$^{1}$
\\
$^{1}$ Mohamed Bin Zayed University of Artificial Intelligence \quad $^{2}$ Alexandria University
\\
{\tt\small \{eman.ali, sathira.silva, muhammad.haris\}@mbzuai.ac.ae}
}

\maketitle

\begin{abstract}
   Vision-language models (VLMs), e.g., CLIP, have shown remarkable potential in zero-shot image classification. However, adapting these models to new domains remains challenging, especially in unsupervised settings where labeled data is unavailable. Recent research has proposed pseudo-labeling approaches to adapt CLIP in an unsupervised manner using unlabeled target data. Nonetheless, these methods struggle due to noisy pseudo-labels resulting from the misalignment between CLIP's visual and textual representations. This study introduces \ours, an unsupervised domain adaptation method for VLMs. \ours introduces the concept of dual prototypes, acting as distinct classifiers, along with the convex combination of their outputs, thereby leading to accurate pseudo-label construction. Next, it ranks pseudo-labels to facilitate robust self-training, particularly during early training. Finally, it addresses visual-textual misalignment by aligning textual prototypes with image prototypes to further improve the adaptation performance. Experiments on 13 downstream vision tasks demonstrate that \ours significantly outperforms zero-shot CLIP and the state-of-the-art unsupervised adaptation baselines.
\end{abstract}

\section{Introduction}
\label{sec:intro}

\noindent\textbf{Context and background:} Vision-language models (VLMs)~\cite{clip,jia2021scaling, yang2022vision, li2022supervision} have shown promising potential in zero-shot image classification. One notable example is CLIP~\cite{clip}. Despite the impressive zero-shot capabilities of CLIP, its performance can be impacted by the discrepancy between the pretraining image-text pairs and the downstream task images~\cite{clip, an2024perceptionclip}. To address this limitation, several studies have attempted to enhance CLIP transfer performance on downstream tasks by leveraging limited labeled samples from the target domains~\cite{zhang2022tip, zhou2022learning, zhou2022conditional, khattakMaPLe, khattak2023selfregulating}. However, in many practical applications, such as security and medical diagnostics, collecting labeled samples can be particularly challenging due to high labeling costs and data privacy concerns. In such contexts, unsupervised learning presents a promising alternative.

Several contributions for adapting CLIP to a target domain using an unlabeled dataset have recently been introduced~\cite{li2022masked, lafter, pouf, upl, inmap, reclip}. A common approach involves leveraging CLIP to generate pseudo-labels, which are then used to fine-tune CLIP in an unsupervised manner. However, this method encounters significant challenges due to the noisy pseudo-labels generated by CLIP. Despite efforts by recent methods for the unsupervised adaptation of CLIP, a significant modality gap persists between the text and vision representations~\cite{reclip, liang2022mind}, leading to inaccurate pseudo-labels causing confirmation bias during adaptation.

\begin{figure}[ht]
  \centering
  \includegraphics[width=0.8\linewidth]{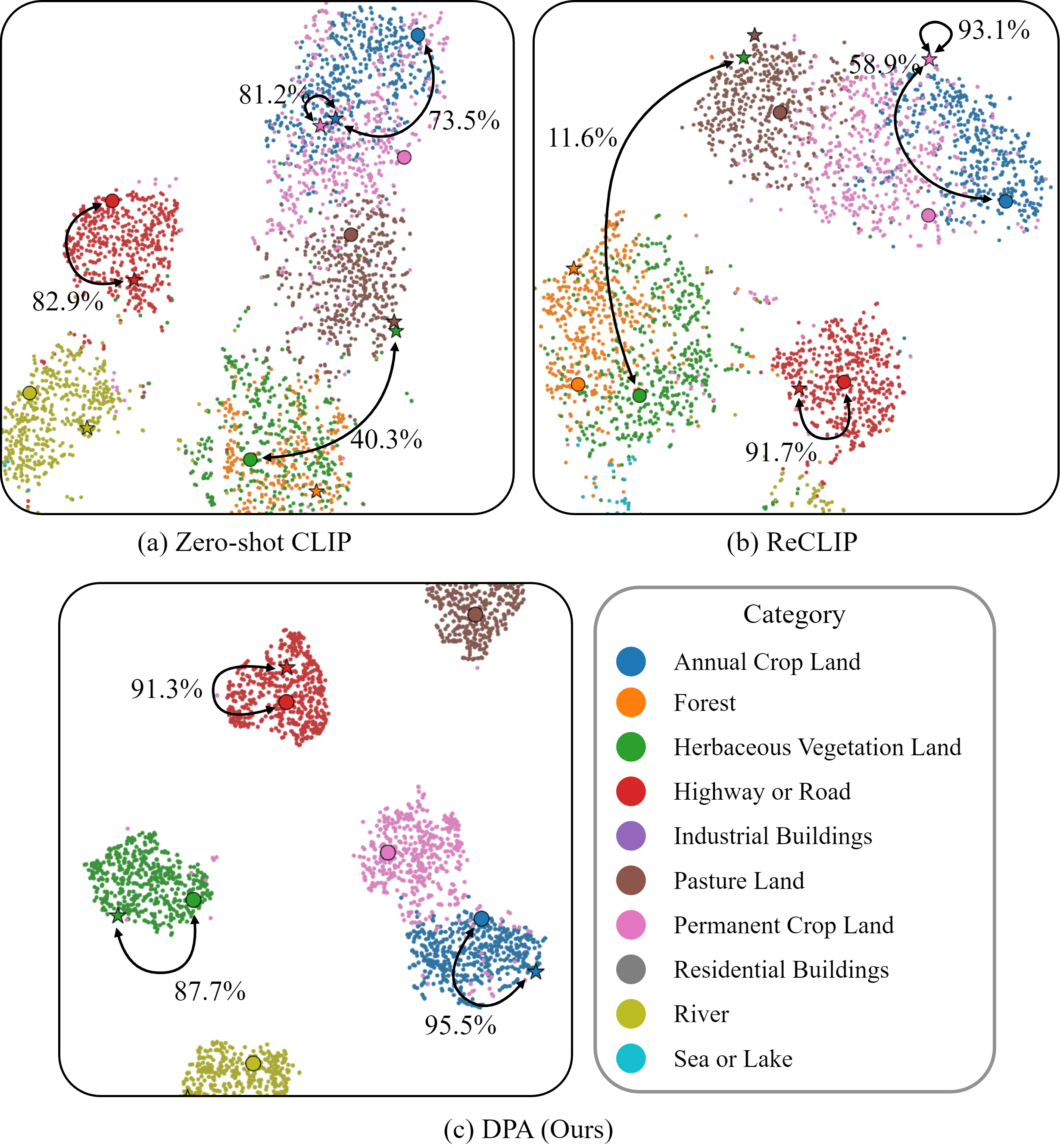}
  \caption{Comparison of t-SNE projections for zero-shot CLIP~\cite{clip}, ReCLIP~\cite{reclip}, and \ours visual embeddings, along with their corresponding visual (circle $\newmoon$) and textual (star $\bigstar$) prototypes on the EuroSAT dataset. The visual prototypes are computed as the mean of each cluster. For clarity, class-agnostic and redundant features are removed from all embeddings using a fixed projection prior to applying the t-SNE projection, following~\cite{reclip}. The cosine similarities between visual and textual prototypes and the well-separated visual clusters, as illustrated in (c), demonstrate the superior performance of our method in capturing both inter-modal and intra-modal alignment compared to the existing approaches depicted in (a) and (b).}
  \label{fig:motivation}
\end{figure}

\noindent\textbf{Motivation:} Recent studies highlight a significant factor behind the performance issues of CLIP in unsupervised adaptation scenarios: the visual domain gap between the source images used to train CLIP and the target images typically occurs when the target samples originate from an uncommon domain~\cite{reclip}. As shown in \cref {fig:motivation}-(a), the t-SNE projection of visual and textual embeddings on the EuroSAT dataset~\cite{helber2019eurosat} reveals a significant misalignment, resulting in misclassifications. Contemporary methods (e.g.,~\cite{reclip}) attempt to address this misalignment by learning a projection space to mitigate the issue and using label propagation~\cite{Iscen_2019_CVPR} to improve pseudo-labels. However, they could be limited as they are either computationally expensive and/or show suboptimal performance in an inductive setting. Further, on datasets with a large number of classes, they struggle to perform well, necessitating turning off label propagation and using model predictions.  
Effectively addressing the challenges of unsupervised adaptation of CLIP in an inductive setting is crucial for improving CLIP's performance in unsupervised adaptation scenarios.

\begin{figure*}
    \centering
    \includegraphics[width=0.95\textwidth, keepaspectratio=true, scale=0.75]{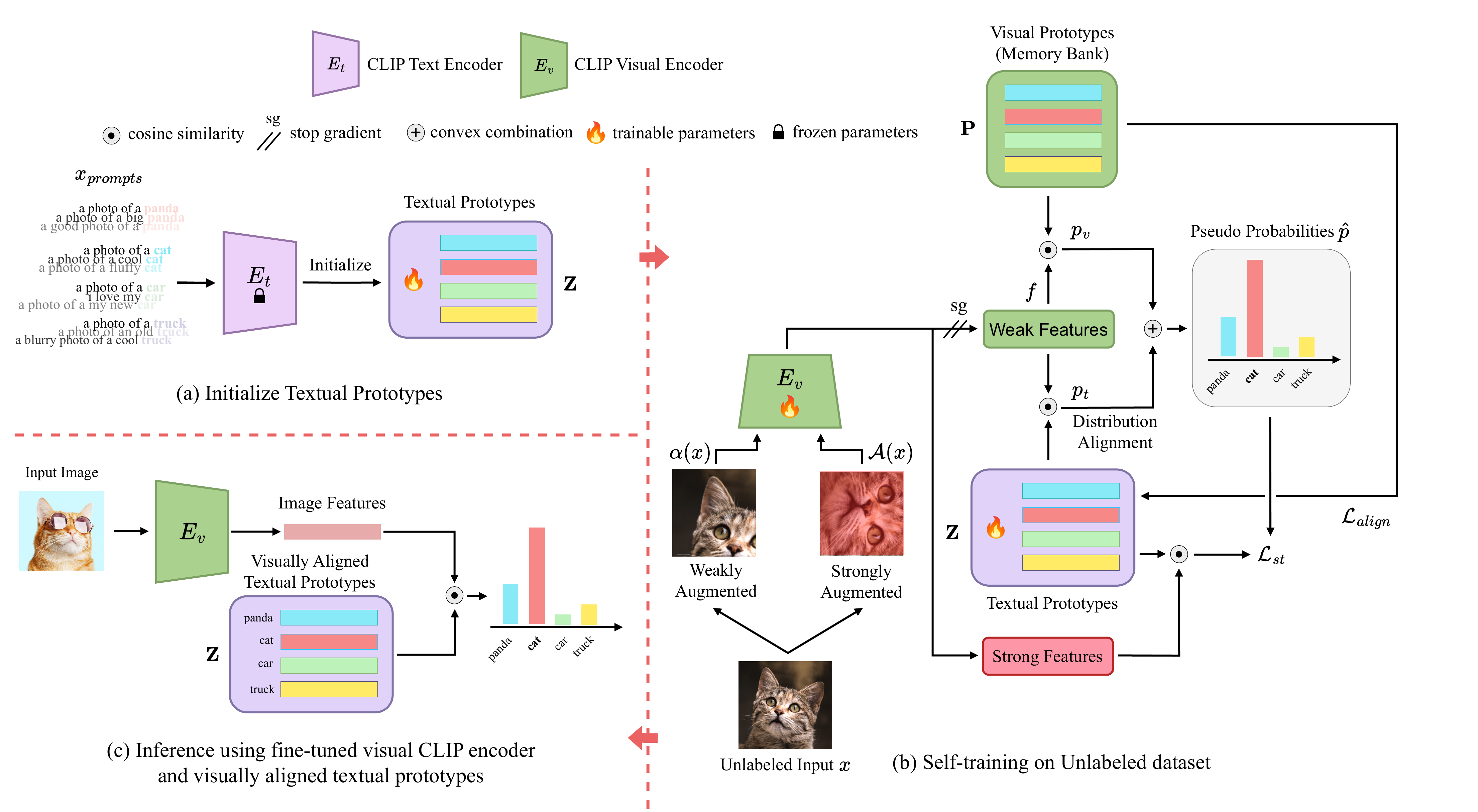}
    \caption{
    The overall framework of \ours.  
    (a) Given a target dataset, \ours utilizes a set of carefully designed prompts to initialize the textual prototypes using the CLIP's zero-shot textual encoder $E_t$. (b) To achieve effective self-training, \ours introduces dual prototypes, namely image and textual prototypes, that behave like two distinct classifiers, and it fuses their outputs via convex combination to form accurate pseudo-labels (PLs). Moreover, it ranks PLs for classification loss to alleviate noisy PLs impact during early self-training. Finally, \ours aligns textual and visual prototypes to adeptly adjust to the target feature semantic relations learned by the visual encoder. (c) During inference, \ours discards the visual prototypes and relies solely on the textual prototypes for prediction.}
    \label{fig:main_method}
     \vspace{-1.5em}
\end{figure*}

\noindent\textbf{Our proposal:} To address these challenges, we propose \ours, a novel unsupervised adaptation method for VLMs (\Cref{fig:main_method}). \ours aims at addressing the domain gap between visual and textual representations in downstream tasks. \ours introduces the idea of dual prototypes, namely image and textual prototypes, which act as distinct classifiers, and their outputs are fused via a convex combination towards generating accurate pseudo-labels. 
There are two main reasons for introducing dual prototypes.
Firstly, the misalignment between the image representation and its textual representation in zero-shot CLIP, due to domain shift, often leads to inaccurate pseudo-labels (\Cref{fig:motivation}-(a)). Secondly, image prototypes, which tend to be less affected by noise, as shown in \cref{fig:motivation}, are generally closer to the true image representation than textual prototypes. Additionally, we tackle the challenge of misalignment between visual and textual embeddings by aligning textual prototypes with image prototypes. This alignment process further enhances the performance of unsupervised CLIP adaptation.

\noindent\textbf{Contributions:}
\textbf{1)} We propose a novel dual-modality prototypic alignment framework, namely \ours for adapting VLMs in an unsupervised manner. Our approach introduces a dual-prototype pseudo-labeling strategy, where the visual prototypes serve as non-parametric classifiers to bring target visual embeddings closer to their respective class prototypes. The learnable textual prototypes, initialized by ensembling different prompt templates using zero-shot CLIP, provide rich task-specific semantics that CLIP visual embeddings typically lack. 
\textbf{2)} We propose ranking pseudo-labels in the classification loss to mitigate noise, particularly during early training. Furthermore, we address visual-textual misalignment by aligning textual prototypes with image prototypes, ensuring the separation of different semantics while aligning with relevant ones.
\textbf{3)} Extensive experiments on 13 downstream vision tasks demonstrate consistent and significant performance enhancements over zero-shot CLIP and the state-of-the-art baselines.

\section{Related Work}
\label{related_work}

\noindent\textbf{Large-scale Vision-Language (VL) Models:} Among other vision-language models \cite{desai2021virtex, sariyildiz2020learning, jia2021scaling, clip, cui2022contrastive, yang2022vision, hu2022scaling}, CLIP~\cite{clip} stands out as a pioneering example, aligning visual and textual features via a contrastive objective on a vast web-crawled collection of image-text pairs. This alignment empowers CLIP to generalize to diverse downstream classification tasks effectively. Building on CLIP's success, subsequent research has explored methods for efficiently transferring the pre-trained model to handle diverse downstream tasks with limited labeled target data. These approaches often leverage vision-specific adapters~\cite{gao2024clip,zhang2022tip, zhang2023prompt, zhu2023not}, language-specific adapters~\cite{zhou2022learning, zhou2022conditional, khattak2023selfregulating}, or both~\cite{khattakMaPLe}. However, these techniques require a minimum number of labeled samples, which poses challenges due to data scarcity, annotation expenses, privacy issues, and practical limitations. Recent advancements have seen the emergence of unsupervised adaptation techniques, focusing on tailoring CLIP to target tasks using unlabeled datasets~\cite{upl, pouf, reclip, lafter, li2022masked, inmap}. However, these methods still depend on additional supervision signals, such as fine-tuning classifiers with a large language model such as GPT-3, as seen in LaFTer~\cite{lafter}. They also often require significant computational resources, as exemplified by MUST~\cite{li2022masked}. 
ReCLIP~\cite{reclip} tackles misaligned embeddings through source-free domain learning, utilizing a projection space to align these embeddings and employing pseudo-labels for self-training. 
Despite progress, there remains significant potential to improve the performance of unsupervised adaptation techniques across various image classification benchmarks.
In contrast, our work introduces a fully label-free approach for adapting VLMs to a target task by generating more accurate pseudo-labels grounded on image and text prototypes.

\noindent\textbf{Prototype-based Learning:}
Prototype-based learning uses a fixed set of distinctive prototypes representing the data, and then the learning is achieved by comparing the test samples directly with these prototypes. Prototype-based learning has been extensively studied in various contexts, including few-shot classification~\cite{snell2017prototypical, liang2022few}, unsupervised learning~\cite{wu2018unsupervised, xu2020attribute} and supervised classification~\cite{yang2018robust, mettes2019hyperspherical}.
Our method innovatively incorporates two types of prototypes, image prototypes, and textual prototypes, to harness both prototypes to ultimately generate accurate pseudo-labels and effectively adapt CLIP to the target dataset.

\noindent\textbf{Pseudo-labeling:} 
Pseudo-labeling is popular in various domains, including vision~\cite{zhai2022lit, sahito2022better} and vision-language tasks~\cite{li2022masked, lafter, upl, reclip}. The pseudo-labeling pipeline of \ours leverages consistency regularization~\cite{sohn2020fixmatch} to produce consistent predictions across different data augmentations, thereby enhancing pseudo-labeling accuracy.
\section{Methodology}
\label{methodology}
This study addresses the task of unsupervised adaptation of vision-language models (i.e., CLIP) for image classification. 
Pre-trained CLIP consists of visual encoder \(E_v\) and textual encoder \(E_t\). The target dataset \(\mathcal{D}_t=\left\{\mathcal{X}_t\right\}\) consists of unlabeled images \(\left\{x_i\right\}_{i=1}^N\), where \(x_i \in \mathcal{X}_t\), and we have unique class names \(\mathcal{C}\) \(=\{c_j\}_{j=1}^C\) for the unlabeled target dataset. 
We use a pre-trained CLIP to process the unlabeled target data \(\mathcal{D}_t\). Initially, the source model assigns a pseudo-label to each unlabeled target image \(x_i\). During the adaptation phase, we employ two prototypes to generate pseudo-labels for adapting CLIP to the target domain.

\subsection{Zero-shot Visual Classification in CLIP}
CLIP achieves impressive zero-shot performance on image classification tasks by learning a joint embedding space that aligns image and text representations. During pre-training, CLIP minimizes a symmetric contrastive loss between semantically similar image-text pairs \({\left(x_i^s, \mathbf{t}_i^s\right)}_{i=1}^n\) sampled from a large-scale source dataset \(\mathcal{D}_s\): 
\begin{equation}
    \sum_i -\log\frac{\exp(f_i^s{}^T \cdot {z}_i / \tau)}{\sum_j^m\exp(f_i^s{}^T\cdot {z}_j / \tau)} -\log\frac{\exp({z}_i^T \cdot f_i^s / \tau)}{\sum_j^m\exp({z}_j^T \cdot f_i^s / \tau)},
\label{eq:clip}
\end{equation}
where \( f_i^s = E_v(x_i^s) \) and \(z_i = E_t(\mathbf{t}^s_i)\), with both \(f_i^s \in \mathbb{R}^{1 \times d}\) and \(z_i \in \mathbb{R}^{1 \times d}\). The parameter \(\tau\) represents a learned temperature, while \(m\) indicates the mini-batch size.
For inference, given an unlabeled target image \(x\)
and a list of class names \(\mathcal{C}\), we leverage the pre-trained textual encoder \(E_t\) to generate text embeddings for each class. Specifically, a series of well-crafted prompts are fed into \(E_t\), producing a set of text embeddings for each class, denoted as \(\mathbf{Z}\)  
where \(\mathbf{Z} \in \mathbb{R}^{C \times d}\). The zero-shot prediction $\hat{y}$ is then obtained as follows:
\begin{equation}
   \hat{y} = \arg\max_{c} (f \cdot \mathbf{Z}^T) 
\label{eq:pseduo}
\end{equation}

\subsection{Proposed Framework (\ours)}
\label{sec:framework}
\noindent\textbf{Motivation:} While zero-shot CLIP shows impressive performance across diverse domains, it still exhibits notable limitations compared to models adapted under supervised conditions due to the misalignment between image and corresponding textual representations~\cite{clip}. As depicted in \cref{fig:motivation}-(a), this discrepancy is evident in CLIP when applied to uncommon domains. Such disparities can lead to inaccurate pseudo-labels, especially when utilizing zero-shot CLIP to generate pseudo-labels for the unsupervised adaptation~\cite{reclip}.
Our objective is to adapt CLIP to the target domain using its unlabeled samples by bridging the gap between image representations and their textual representations, all without dependence on ground-truth labels. Simultaneously, we seek to achieve efficient parameter-based unsupervised adaptation for CLIP. Our objective relies on two key components: (1) ensuring the accuracy of the pseudo-labels generated for the unlabeled samples and (2) aligning textual representations with their corresponding visual representations.
To generate the pseudo-labels for adapting CLIP to the target domain, we introduce dual prototypes that act as two different classifiers and fuse their complementary outputs via a convex combination. Furthermore, we propose visual-textual prototype alignment towards further improving the adaptation performance.

\noindent\textbf{Textual Prototypes Construction:}
To create textual prototypes, we obtain the textual representation for each class in the target domain. This is achieved by incorporating each class name \(c_j\) into a predefined prompt template. It is then processed by CLIP's text encoder to produce a textual representation \(z_j\). We utilize \(k\) distinct prompts for each class, resulting in \(k\) corresponding textual representations \(z_j \in \mathbb{R}^{k \times d}\), where \(d\) is the dimensionality of the embedding space. The textual prototype \(\mathbf{Z}_j\) for class \(c_j\) is constructed by averaging these \(k\) individual textual representations:
\begin{equation}
    \mathbf{Z}_j = \frac{1}{k} \sum^{k}_{i=1}{z_{ji}} 
\label{eq:text_prototypes}
\end{equation}
where \(\mathbf{Z}_j \in \mathbb{R}^{1 \times d}\). Finally, we define the textual prototypes for the set of \(C\) classes as \(\mathbf{Z} \in \mathbb{R}^{C \times d}\). This approach enables us to capture a comprehensive and robust representation of class semantics by consolidating textual representations derived from multiple prompts. Finally, these textual prototypes act as the initialization for a parametric classifier that is updated during training alongside CLIP parameters. 

\noindent\textbf{Image Prototypes Construction:}
\label{sec:image_proto}
To construct the image prototypes, CLIP's image encoder processes a weakly-augmented unlabeled image \(\alpha(x)\) and generates a feature representation \(f = E_v(\alpha (x))\), where \(f \in \mathbb{R}^{1 \times d}\). Without class labels, we utilize \cref {eq:clip} and \cref{eq:pseduo} to generate pseudo-labels \(\hat{y}\) from zero-shot CLIP. 
To obtain the image prototype \(\mathbf{P}_j\) for a class \(c_j\), we average the feature representation \(f\) of the unlabeled images assigned to class \(c_j\) as follows:
\begin{equation}
    \mathbf{P}_j = \frac{1}{N_j} \sum_{i=1}^{N} \!\mathbb{I}\{\hat{y_i}=j \!\}\! \; f_i
\label{eq:img_proto}
\end{equation}   
where \(N_j =  \sum_{i=1}^{N} \!\mathbb{I}\{\hat{y_i}=j \!\}\! \). For the set of \(C\) classes, we define the image prototypes as \(\mathbf{P} \in \mathbb{R}^{C \times d}\)—the image prototypes function as a non-parametric classifier.
Given that the pseudo-labels generated by zero-shot CLIP are initially noisy, our approach, inspired by previous works~\cite{zhou2024source, wen2016discriminative, xie2018learning}, employs a memory bank \(\mathbf{MB}\). This memory bank facilitates the incremental update of image prototypes in a non-parametric way throughout the training process. Specifically, \(\mathbf{MB}\) stores the feature representations of all unlabeled target samples along with their corresponding pseudo-labels: \(\mathbf{MB} = \{(f_i, \hat{y}_i)\}_{i=1}^N\). As the image prototypes are non-parametric, we iteratively refine them using the memory bank as follows: (1) We initially calculate the image prototypes using \cref{eq:img_proto}, utilizing the image features and the pseudo-labels generated from zero-shot CLIP across the entire target dataset. (2) During training, we continuously update the memory bank by storing the image representations alongside their corresponding pseudo-labels. (3) At the end of each epoch, we update the prototypes \(\mathbf{P}_j\) using \cref{eq:img_proto}, incorporating the image representations and their corresponding pseudo-labels from \(\mathbf{MB}\). (4) We repeat steps (2) and (3) until the completion of the training process.
\ours maximizes the utilization of all available training samples to enhance the image prototypes. This iterative process ensures that each prototype continuously evolves, progressively assimilating relevant knowledge from individual training samples throughout the model training phase while mitigating the error propagation commonly associated with pseudo-labels.

\noindent\textbf{Pseudo-labels Generation:} 
\label{sec:refine}
Both image and textual prototypes serve as distinct classifiers for the unlabeled target dataset. We generate a pseudo-label (PL) for the weakly-augmented unlabeled target image \(\alpha (x)\) from these two prototypes. Initially, we determine the similarity between the image features and the textual prototypes as follows: 
\begin{equation}
p_t = f \cdot \mathbf{Z}^T
\label{eq:pseduo_text}
\end{equation}
Then, we find the similarity between the image features and the image prototypes as follows:
\begin{equation}
p_v= f \cdot \mathbf{P}^T
\label{eq:pseduo_img}
\end{equation}
Finally, we fuse \(p_t\) and \(p_v\) to form \(\hat{p}\) as follows: 
\begin{equation}
\label{eq:da}
    \hat{p} = \beta \cdot DA(p_t) + (1 - \beta) \cdot p_v
\end{equation}
Following~\cite{li2021comatch}, we perform distribution alignment (DA) to prevent the model’s prediction from collapsing onto specific classes. Here, \(DA(p_t) = p_t / \bar{p}_t\), where \(\bar{p}_t\) represents a running average of \(p_t\) during training, and \(\beta\) is a hyperparameter. Finally, the PL is obtained as \(\hat{y} = \arg\max_{c} \hat{p}\).
This process of generating pseudo-labels improves the accuracy of the PL during training, indicating that these PLs may initially be noisy (see \cref{fig:ours_vs_base}). Therefore, assigning uniform weights to all pseudo-labels could hinder the adaptation process. To address this challenge, we propose adjusting the classification loss weight for the pseudo-label \(\hat{y}\) based on its similarity to the textual and image prototypes, as follows: 
\begin{equation}
\label{eq:reweighting}
w_{x} = \langle f, \mathbf{Z}(\hat{y}) \rangle \langle f, \mathbf{P}(\hat{y}) \rangle
\end{equation}
\(\langle \cdot,\cdot \rangle\) represents the cosine similarity between the image features and the corresponding prototypes derived from the PLs.
Finally, we utilize the pseudo-label $\hat{y}$, obtained from a weakly-augmented image $\alpha (x)$, as self-supervision for the strongly-augmented counterpart \(\mathcal{A}(x)\) as follows: 
\begin{equation}
    \label{eq:self-training}
    \mathcal{L}_{st} = - \; \mathbb{E}_{x \in \mathcal{X}_t} \; \left[ w_{x} \cdot \sum_{j=1}^C  \!\mathbb{I}\{\hat{y}=j \!\}\! \; log \left(p_{\mathcal{A}(x)}\right) \right],
    \end{equation}
where \(w_{x}\) represents the weight calculated from \cref{eq:reweighting}, and \(p_{\mathcal{A}(x)} = E_v(\mathcal{A}(x)) \cdot \mathbf{Z}^T\) represents the probabilistic output for the strongly-augmented image \(\mathcal{A}(x)\). To alleviate the confirmation bias induced by CLIP~\cite{wang2022debiased}, we adopt “fairness” regularization as suggested in~\cite{li2022masked} as follows: 
\begin{equation}
\label{eq:reg}
   \mathcal{L}_{reg} = -\frac{1}{C} \sum_{j=1}^C \log\left(\bar{p}_{\mathcal{A}(x_j)}\right),
\end{equation}    
where \(\bar{p}_{\mathcal{A}(x)}\) represents the model's average prediction from the strongly augmented images across the batch.
By incorporating this fairness regularization, we aim to encourage the model to make more uniform predictions across classes, reducing the tendency to overfit the PLs and promoting a more balanced adaptation to the target domain. \cref{eq:self-training} and \cref{eq:reg} are used to update both the image encoder and the textual prototypes. 

\noindent\textbf{Prototypes Alignment:}
\label{sec:align}
During training, the textual prototypes undergo continual updates to align with the associated image representation, leveraging pseudo-labels using \cref{eq:self-training} and \cref{eq:reg}. Conversely, the refinement of the image prototypes occurs through averaging image features, as detailed in Section~\ref{sec:image_proto}. 
We exploit the image prototypes to provide additional supervision for updating textual prototypes beyond pseudo-label alone. These image prototypes, less influenced by erroneous pseudo-labels and closer to image features (as illustrated in \cref{fig:motivation}), offer extra guidance for refining textual prototypes. Our primary objective is to optimize the relationship between image prototypes \(\mathbf{P}\) and textual prototypes \(\mathbf{Z}\) by maximizing the (cosine) similarity between these feature sets.
To accomplish this optimization, we employ the InfoNCE loss~\cite{oord2018representation} to directly assess the alignment between the corresponding feature vectors in \(\mathbf{P}\) and \(\mathbf{Z}\), as follows:
\begin{equation}
    \label{eq:align}
    \mathcal{L}_{\text{align}} = -\sum_{j=1}^C \log \dfrac{\exp(\mathbf{P}_j \cdot \mathbf{Z}_j^T / \tau)}
    {\displaystyle \sum_r \exp(\mathbf{P}_j \cdot \mathbf{Z}_r^T / \tau)}
\end{equation}
With this alignment, we maximize the (cosine) similarity between the image prototypes \(\mathbf{P}\) and the textual prototypes \(\mathbf{Z}\), encouraging the model to learn more closely aligned representations.
The overall loss function used for training on the target data is, \(\mathcal{L} = \lambda_1 \mathcal{L}_{st} + \lambda_2 \mathcal{L}_{reg} + \lambda_3 \mathcal{L}_{align}\), 
where \(\lambda_1\), \(\lambda_2\), \(\lambda_3\) are non-learnable parameters.

\begin{table*}[!ht]
  \centering
  \resizebox{\textwidth}{!}{
    \begin{tabular}{@{}lcccccccccccccc@{}}
      \toprule
      \textbf{Method} & \textbf{ImgNet} & \textbf{Caltech} & \textbf{DTD} & \textbf{ESAT} & \textbf{FGVCA} & \textbf{Food} & \textbf{Flower} & \textbf{OxPets} & \textbf{SUN} & \textbf{StCars} & \textbf{CIFAR10} & \textbf{CIFAR100} & \textbf{UCF} & \textbf{Avg} \\
      \midrule
      Zero-shot CLIP~\cite{clip} & \underline{63.30} & 90.69 & 44.42 & 43.84 & 19.50 & 82.40 & 66.46 & 87.50 & 61.99 & \underline{58.74} & 89.80 & 65.10 & 64.20 & 64.46 \\  
      UPL~\cite{upl}   & 58.22 & 92.36 & 45.37 & 51.88 & 17.07 & \underline{84.25} & 67.40 & 83.84 & 62.12 & 49.41 & 91.26 & 67.41 & 62.04 & 64.05 \\  
      POUF~\cite{pouf} & 52.20 & 94.10 & 46.10 & 62.90 & 18.20 & 82.10 & 67.80 & \underline{87.80} & 60.00 & 57.70 & 90.50 & 62.00 & 61.20 & 64.82 \\  
      LaFTer~\cite{lafter} & 61.63 & \underline{94.39} & \underline{50.32} & \underline{69.96} & \underline{19.86} & 82.45 & \underline{72.43} & 84.93 & \underline{65.87} & 57.44 & \underline{94.57} & \underline{69.79} & \underline{65.08} & \underline{68.36} \\  
      \textbf{\ours} & \textbf{64.64} & \textbf{96.06} & \textbf{55.69} & \textbf{80.04} & \textbf{20.67} & \textbf{84.76} & \textbf{75.56} & \textbf{90.71} & \textbf{68.13} & \textbf{62.62} & \textbf{95.97} & \textbf{76.47} & \textbf{68.49} & \textbf{72.29} \\
      \bottomrule
    \end{tabular}
  }
  \caption{Comparison of state-of-the-Art unsupervised adaptation methods using the ViT-B/32 backbone.}
  \label{table:sota_vit_b_32}
\end{table*}
\section{Experiments}
\noindent\textbf{Datasets and Baselines:} We extensively evaluate on 13 diverse datasets: ImageNet~\cite{deng2009imagenet}, Caltech101~\cite{fei2004learning}, DTD~\cite{cimpoi2014describing}, EuroSAT~\cite{helber2019eurosat}, FGVCAircraft~\cite{maji2013fine}, Food101~\cite{bossard2014food}, Flowers102~\cite{nilsback2008automated}, OxfordPets~\cite{parkhi2012cats}, SUN397~\cite{xiao2010sun}, StandfordCars~\cite{krause20133d}, CIFAR10\textbackslash100~\cite{krizhevsky2009learning}, and UCF101~\cite{soomro2012ucf101}.
We perform comparative analysis with four state-of-the-art (SOTA) unsupervised adaptation methods: CLIP~\cite{clip}, UPL~\cite{upl}, POUF~\cite{pouf}, and LaFTer~\cite{lafter}.

\noindent\textbf{Implementation Details:} 
For all experiments, unless otherwise specified, we utilize a ViT/B-32 CLIP pre-trained by OpenAI~\cite{clip}. In the context of unsupervised fine-tuning, we specifically target the layer-normalization weights of the image encoder and the textual prototypes. This strategy is effective and stable for adapting models with noisy supervision~\cite{ba2016layer,wang2020tent}. The text encoder of CLIP will be excluded after constructing the textual prototypes. 
Input images are standardized to a size of 224 × 224. During training, we apply a series of data augmentation techniques, including RandomResizedCrop, Flip, RandAugment~\cite{cubuk2020randaugment} as a strong augmentation and Resize+RandomCrop for generating pseudo-labels. For testing, we extract a Center crop after resizing the image to $224$.
We utilize AdamW optimizer~\cite{loshchilov2017decoupled} with a cosine learning rate schedule. To ensure fair comparisons, we reproduced the results of UPL, POUF, and LaFTer using their publicly available code. Most state-of-the-art (SOTA) methods use different backbones, and UPL's reported results were only for 16 shots, unlike our approach, which provides results for the entire training dataset.
For additional details on the implementation, see the supplementary materials.

\noindent\textbf{Results:} \Cref{table:sota_vit_b_32} shows that 
our approach consistently outperforms zero-shot CLIP, improving top-1 accuracy by $+7.83\%$ on average. Compared to UPL, which optimizes prompts, \ours still achieves consistent improvements with an average gain of $+8.24\%$ with minimal additional computational overhead despite slightly larger parameters. We observe that UPL's performance on the entire unlabeled training dataset is inferior to that of zero-shot CLIP. This discrepancy may stem from the fact that UPL generates offline pseudo-labels using zero-shot CLIP, which can lack confidence, particularly when dealing with fine-grained classes in certain datasets~\cite{lafter}. 
\ours also surpasses POUF and LaFTer, achieving average increases of $+7.47\%$ and $+3.93\%$, respectively, across the 13 datasets without requiring additional text corpora or further pre-training. We note significantly larger performance improvement on the ESAT dataset compared to other datasets. We believe this is because zero-shot CLIP exhibits limitations when applied to specialized, complex, or abstract tasks, such as satellite image classification~\cite{clip}. This suggests there is potential for improvement through additional supervision. This behavior aligns with findings from previous studies. LaFTer achieved a remarkable +26.12\% improvement over zero-shot CLIP.
ReCLIP reported gains exceeding +26.96\% (\Cref{table:reclip}). The 10\% improvement by our method (\ours) over LaFTer highlights the effectiveness of our pseudo-labeling approach. 

\noindent \textbf{For Ablation Studies}, we use 11 out of 13 datasets, excluding ImageNet and SUN397 due to their large size. 
Excluding these large-scale datasets allows us to perform more extensive experiments and analyses while maintaining computational feasibility.

\begin{figure}[!ht]
  \centering
  \includegraphics[width=0.8\linewidth]{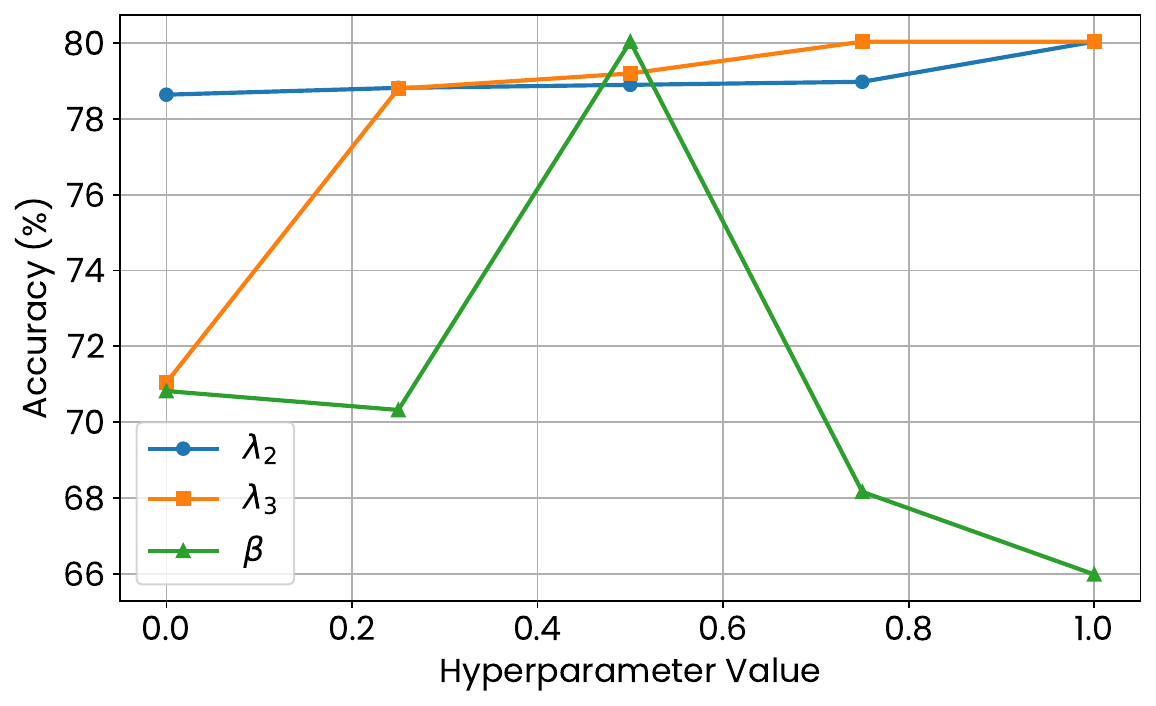}
  \caption{Sensitivity analysis of the hyperparameters $\lambda_2$, $\lambda_3$, and $\beta$ on the accuracy ($\%$) of \ours. We set $\lambda_1=1$ across all datasets (here ESAT is used) since $\mathcal{L}_{st}$ is our primary loss.}
  \label{fig:sensitivity}
\end{figure}

\noindent\textbf{Sensitivity Analyses of Hyperparameters:} The sensitivity analyses of the hyperparameters $\lambda_2$, $\lambda_3$, and $\beta$ on the ESAT dataset are presented in \cref{fig:sensitivity}.

\noindent \textbf{Results on Different Backbones}
We compare different backbones (ViT-B/16) in \cref{table:sota_vit_b_16} \footnote{A fair comparison using RN50
is not possible, as our relevant baselines were specifically designed for the ViT image encoder architecture (For example, LaFTer~\cite{lafter} / ReCLIP~\cite{reclip})}.

\begin{table}[!htp]
  \centering
  \resizebox{\columnwidth}{!}{%
    \begin{tabular}{@{}lcccccccccccc@{}}
      \toprule
      \textbf{Method} & \textbf{Caltech} & \textbf{DTD} & \textbf{ESAT} & \textbf{FGVCA} & \textbf{Flower} & \textbf{OxPets} & \textbf{StCars} & \textbf{CIFAR10} & \textbf{CIFAR100} & \textbf{UCF} & \textbf{Avg} \\ 
      \midrule
      Zero-shot CLIP & 92.60 & 44.70 & 49.00 & 23.97 & 70.89 & 89.00 & 64.70 & 90.80 & 68.22 & 69.10 & 66.30 \\  
      UPL            & 95.14 & 45.90 & 55.36 & 22.53 & 73.93 & 87.98 & 60.33 & 92.92 & 70.41 & 67.43 & 67.19 \\  
      POUF           & 95.40 & 48.60 & 59.50 & \underline{24.40} & 72.10 & \underline{91.80} & 63.50 & 92.30 & 66.30 & \underline{71.50} & 68.54 \\ 
      LaFTer         & \underline{95.92} & \textbf{54.79} & \underline{72.10} & 22.38 & \underline{75.15} & 85.28 & \underline{64.72} & \textbf{96.50} & \underline{76.30} & 67.20 & \underline{71.03} \\  
      \textbf{\ours} & \textbf{96.09} & \underline{50.69} & \textbf{81.22} & \textbf{25.14} & \textbf{78.68} & \textbf{93.19} & \textbf{68.45} & \underline{96.43} & \textbf{78.10} & \textbf{74.62} & \textbf{74.26} \\  
      \bottomrule
    \end{tabular}%
  }
  \caption{Top-1 accuracy (\%) comparison of \ours with SOTA unsupervised adaptation methods using CLIP-ViT-B/16.}
  \label{table:sota_vit_b_16}
\end{table}

\begin{figure*}[ht]
  \centering
  \includegraphics[width=0.9\textwidth]{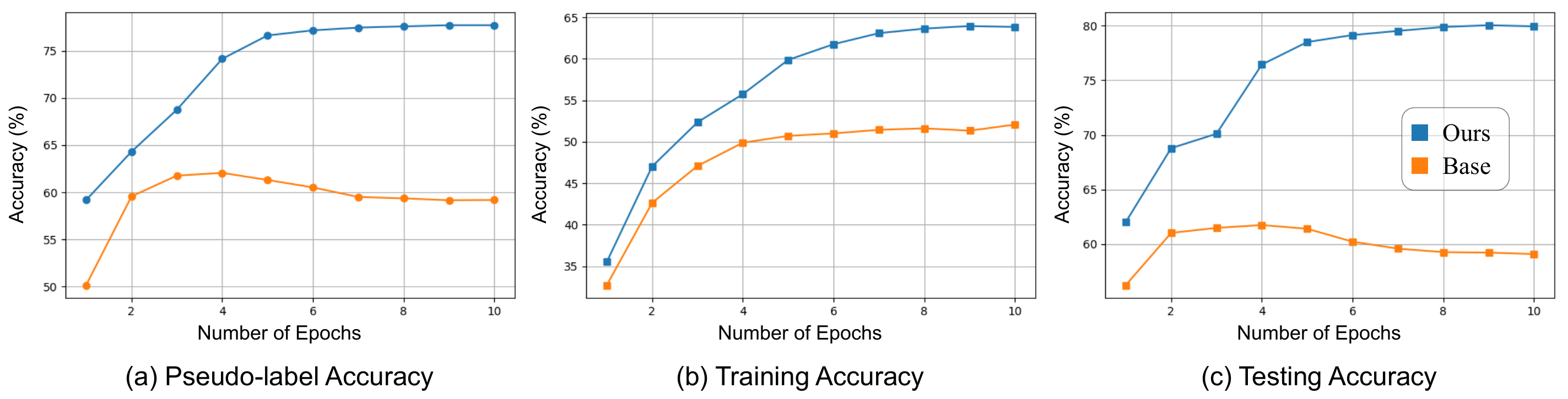}
  \caption{\textbf{Base} vs. \ours in (a) PLs, (b) training, and (c) testing accuracy on EuroSAT \cite{helber2019eurosat} dataset.}
  \label{fig:ours_vs_base}
\end{figure*}

\noindent\textbf{Analyses of Model Components:}  
We investigate the impact of different components through the training of four distinct models:
\textbf{Base}: This model iteratively generates pseudo-labels using textual prototypes and incorporates regularization loss through self-training.
\textbf{Center}: CLIP undergoes training with PLs generation using image and text prototypes, as detailed in Section~\ref{sec:refine}, in addition to building upon the foundation established in the Base model.
\textbf{Center+$w$}: introduces the weighting strategy for \textbf{Center} model.
\textbf{Center+$w$+Align}: introduces the alignment loss, as detailed in Section~\ref{sec:align}. 
As shown in \cref{table:ablation}, \textbf{Base} model, which relies solely on self-training, demonstrates a notable improvement compared to the zero-shot CLIP. However, as training progresses, the impact of confirmation bias becomes evident, leading to a decrease in the accuracy of the pseudo-labels, as illustrated in \cref{fig:ours_vs_base}-(a).
To mitigate the effects of confirmation bias, we introduce the pseudo-label generation strategy in the ~\textbf{Center} model. This approach enhances the quality of pseudo-labels throughout the training process, leading to a significant gain of $+7.28\%$ over zero-shot CLIP in average accuracy.
Incorporating the weighting strategy in \textbf{Center+$w$} model results in an additional increase of $+8.01\%$ over zero-shot CLIP in accuracy. By assigning appropriate weights to the self-training loss, the model can better balance the contributions of high-confidence and low-confidence pseudo-labels, leading to more stable and practical training.
Finally, including the alignment loss in \textbf{Center+$w$+Align} model leads to a further increase of $+8.58\%$ over zero-shot CLIP in average accuracy. By explicitly encouraging the alignment between image and text prototypes, the model learns more robust and transferable representations, which are crucial for effective unsupervised adaptation.
Finally, we show a detailed analysis of the evolution of pseudo-label accuracy during training epochs for our method (\ours) compared to \textbf{Base} model in \cref{fig:ours_vs_base}-a. See more discussions in suppl.

\begin{table*}[!ht]
  \centering
  \resizebox{\textwidth}{!}{
    \begin{tabular}{@{}lcccccccccccc@{}}
      \toprule
      \textbf{Method} & \textbf{Caltech} & \textbf{DTD} & \textbf{ESAT} & \textbf{FGVCA} & \textbf{Food} & \textbf{Flower} & \textbf{OxPets} & \textbf{StCars} & \textbf{CIFAR10} & \textbf{CIFAR100} & \textbf{UCF} & \textbf{Avg} \\
      \midrule
      \textbf{Zero-shot CLIP} & 90.69 & 44.42 & 43.84 & 19.50 & 82.40 & 66.46 & 87.50 & 58.74 & 89.80 & 65.10 & 64.20 & 64.79 \\
      \textbf{Base} & 93.57 & 48.99 & 61.94 & 19.20 & 84.10 & 68.45 & 90.24 & 59.25 & 95.95 & 73.55 & 65.45 & 69.15 \\  
      \textbf{Center} & 95.44 & \underline{55.53} & 70.56 & \underline{19.80} & \underline{84.65} & 75.27 & \textbf{90.71} & \underline{61.53} & 95.96 & \underline{76.01} & 67.30 & 72.07 \\    
      \textbf{Center+$w$} & \underline{95.46} & 54.54 & \textbf{80.06} & 19.56 & 84.63 & \underline{75.44} & \underline{90.49} & 61.19 & \textbf{95.97} & 75.92 & \underline{67.51} & \underline{72.80} \\   
      \textbf{Center+$w$+Align (\ours)} & \textbf{96.06} & \textbf{55.69} & \underline{80.04} & \textbf{20.67} & \textbf{84.76} & \textbf{75.56} & \textbf{90.71} & \textbf{62.62} & \textbf{95.97} & \textbf{76.47} & \textbf{68.49} & \textbf{73.37} \\
      \bottomrule
    \end{tabular}
  }
  \caption{Ablation study. See text for details.}
  \label{table:ablation}
\end{table*}

\noindent\textbf{Incorporating Image Prototypes as Pseudo-Label Source:}
We analyze incorporating image prototypes as an additional classifier. We exclude ~\cref{eq:da} and generate PLs solely based on text prototypes while preserving the alignment of image-text prototypes and the weighting strategy. We update \(\mathbf{MB}\) solely using the pseudo-labels derived from the textual prototypes. ~\Cref{table:no_proto} shows that the alignment and weighting strategies generally perform well across datasets, with minor differences compared to the entire model, except for EuroSAT and Flowers-102. 

\begin{table*}[!ht]
  \centering
  \small
  \resizebox{\textwidth}{!}{%
  \begin{tabular}{@{}lcccccccccccc@{}}
    \toprule
    \textbf{Method} & \textbf{Caltech} & \textbf{DTD} & \textbf{ESAT} & \textbf{FGVCA} & \textbf{Food} & \textbf{Flower} & \textbf{OxPets} & \textbf{StCars} & \textbf{CIFAR10} & \textbf{CIFAR100} & \textbf{UCF} & \textbf{Avg} \\
    \midrule
    \textbf{Zero-shot CLIP} & 90.69 & 44.42 & 43.84 & \underline{19.50} & 82.40 & 66.46 & 87.50 & 58.74 & 89.80 & 65.10 & 64.20 & 64.79 \\
    \textbf{\ours} (w/o PLs Generation) & \underline{95.41} & \underline{54.20} & \underline{61.94} & 19.44 & \underline{84.49} & \underline{69.22} & \textbf{91.17} & \underline{61.95} & \underline{95.93} & \underline{76.12} & \underline{67.96} & \underline{70.71} \\ 
    \textbf{\ours} & \textbf{96.06} & \textbf{55.69} & \textbf{80.04} & \textbf{20.67} & \textbf{84.76} & \textbf{75.56} & \underline{90.71} & \textbf{62.62} & \textbf{95.97} & \textbf{76.47} & \textbf{68.49} & \textbf{73.37} \\
    \bottomrule
  \end{tabular}
  }
  \caption{Image prototypes vs. text prototypes for pseudo-label generation.}
  \label{table:no_proto}
\end{table*}

\noindent\textbf{Impact of Image Prototype Initialization:}
We evaluate three initializations for image prototype methods: mean, weighted mean, and similarity weighted mean~\cite{liang2022few} (\Cref{table:proto_inital}). The results demonstrate significant improvements over the baselines for all initialization strategies.

\begin{table*}[!ht]
  \centering
  \small
  \resizebox{\textwidth}{!}{%
  \begin{tabular}{@{}lcccccccccccc@{}}
    \toprule
    \textbf{Method} & \textbf{Caltech} & \textbf{DTD} & \textbf{ESAT} & \textbf{FGVCA} & \textbf{Food} & \textbf{Flower} & \textbf{OxPets} & \textbf{StCars} & \textbf{CIFAR10} & \textbf{CIFAR100} & \textbf{UCF} & \textbf{Avg} \\
    \midrule
    \textbf{Zero-shot CLIP} & 90.69 & 44.42 & 43.84 & 19.50 & 82.40 & 66.46 & 87.50 & 58.74 & 89.80 & 65.10 & 64.20 & 64.79 \\
    \textbf{\ours} (Weighted Mean) & \underline{96.00} & 54.52 & 68.34 & 20.31 & \underline{84.70} & 73.81 & 90.49 & \textbf{63.14} & \underline{95.96} & 75.81 & 68.04 & 71.92 \\
    \textbf{\ours} (Similarity Weighted Mean) & 94.52 & \underline{55.16} & \underline{79.76} & \textbf{21.12} & \underline{84.70} & \textbf{77.06} & \textbf{90.84} & \underline{62.98} & 95.93 & \underline{76.33} & \underline{68.28} & \underline{73.33} \\
    \textbf{\ours} (Mean) & \textbf{96.06} & \textbf{55.69} & \textbf{80.04} & \underline{20.67} & \textbf{84.76} & \underline{75.56} & \underline{90.71} & 62.62 & \textbf{95.97} & \textbf{76.47} & \textbf{68.49} & \textbf{73.37} \\
    \bottomrule
  \end{tabular}
  }
  \caption{Performance of \ours with three different initializations for image prototypes.}
  \label{table:proto_inital}
\end{table*}

\noindent\textbf{Robustness Analysis under Distribution Shifts:}
To evaluate robustness against natural distribution shifts, we show results on ImageNet-Sketch~\cite{wang2019learning}, ImageNet-Adversarial~\cite{hendrycks2021natural}, and ImageNet-Rendition~\cite{hendrycks2021many}. Initially, we evaluate a fine-tuned ViT-B/32 model on ImageNet using our method, achieving an accuracy of $64.64\%$ on the ImageNet validation set.
We explore transductive transfer learning~\cite{xian2017zero, rohrbach2013transfer} with our approach. Transductive learning involves assuming access to unlabeled test images from the new distribution (\Cref{table:dist_shift}).

\begin{table}[!ht]
  \centering
  \small
  \resizebox{\columnwidth}{!}{%
  \begin{tabular}{@{}lcccc@{}}
    \toprule
    \textbf{Method} & \textbf{ImageNet} & \textbf{ImageNet-S} & \textbf{ImageNet-R} & \textbf{ImageNet-A} \\
    \midrule
    \textbf{Zero-shot CLIP} & 63.30 & \underline{42.10} & \textbf{68.60} & \underline{32.10} \\
    \textbf{CoOp} (M =16) & \textbf{66.85} &  40.44 & 64.45 &  30.62 \\
     \textbf{CoOp} (M =4) & \underline{66.34} &  41.48 & 65.78 &  31.34 \\
    \textbf{\ours} (Transductive) & 64.64 & \textbf{42.60} & \underline{68.40} & \textbf{32.20} \\
    \bottomrule
  \end{tabular}
  }
  \caption{Transductive adaptation to distribution shift.}
  \label{table:dist_shift}
\end{table}

\noindent\textbf{Evaluation of Fine-Tuning Techniques:}
\Cref{table:full} shows the results of various fine-tuning techniques for CLIP, including LoRA~\cite{hu2022lora}, K-Adaptation~\cite{he2023parameter}, and full model fine-tuning, and our approach, which involves fine-tuning only the layer normalization weights (see \cref{table:full}). The last row of the table shows the PLs of zero-shot CLIP used at the start of the training for constructing the initial image prototypes. In datasets characterized by highly noisy PLs, \cref{table:full} underscores the significance of exclusively fine-tuning layer-normalization weights during adaptation. This approach helps to mitigate overfitting and enhances overall performance. Our findings indicate that fine-tuning solely the layer normalization weights is effective and stable for adapting models under noisy supervision.

\begin{table}[!ht]
  \centering
  \resizebox{\columnwidth}{!}{
    \begin{tabular}{@{}lcccccccc@{}}
      \toprule
      \textbf{Method} & \textbf{Caltech} & \textbf{DTD} & \textbf{ESAT} & \textbf{FGVCA} & \textbf{Flower} & \textbf{OxPets} & \textbf{StCars} & \textbf{Avg} \\
      \midrule
      Zero-shot CLIP & 90.69 & 44.42 & 43.84 & 19.50 & 66.46 & 87.50 & 58.74 & 58.74 \\  
      \textbf{\ours} (LoRA~\cite{hu2022lora}) & 90.87 & 50.32 & 65.16 & \underline{19.53} & 66.87 & 87.71 & 57.59 & 62.58 \\
      \textbf{\ours} (K-Adaptation~\cite{he2023parameter}) & 92.49 & 52.50 & \underline{69.80} & 19.38 & 68.09 & 89.04 & \underline{60.81} & \underline{64.59} \\
      \textbf{\ours} (Full Tuning) & \underline{94.57} & \underline{53.03} & 62.92 & 17.37 & \underline{75.23} & \textbf{91.14} & 54.06 & 64.05 \\
      \textbf{\ours} (LN) & \textbf{96.06} & \textbf{55.69} & \textbf{80.04} & \textbf{20.67} & \textbf{75.56} & \underline{90.71} & \textbf{62.62} & \textbf{68.76} \\ 
      \midrule
    \textbf{\ours} (PLs accuracy of Zero-shot CLIP) & 88.83 & 43.54 & 44.90 & 17.79 & 66.53 & 83.80 & 57.81 & 57.6 \\
      \bottomrule
    \end{tabular}
  }
  \caption{Top-1 accuracy (\%) comparison of \ours using different parameter-efficient fine-tuning methods. All experiments are conducted with CLIP-ViT-B/32 as the backbone.}
  \label{table:full}
\end{table}

\begin{figure}[!ht]
  \centering
  \includegraphics[width=0.7\linewidth]{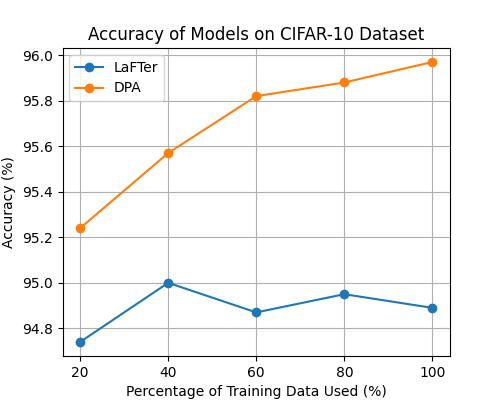}
  \caption{Top-1 accuracy ($\%$) for the ViT-B/32 backbone on the CIFAR-10 dataset using 20$\%$, 40$\%$, 60$\%$, 80$\%$, and 100$\%$ of the training data for \ours. }
  \label{fig:shots}
\end{figure}

\noindent\textbf{Data Scale and Downstream Task Performance:}
We conducted an experiment using the CIFAR-10 dataset to evaluate the scalability of our method (\ours), compared to LaFTer, as illustrated in \cref{fig:shots}. To assess the robustness of our approach, we tested various portions of the dataset: 20\%, 40\%, 60\%, and 80\%. Our findings indicate that the \ours method consistently performs well across different dataset sizes, demonstrating its scalability regardless of the
amount of data used.

\begin{figure}[!ht]
  \centering
  \includegraphics[width=0.8\linewidth]{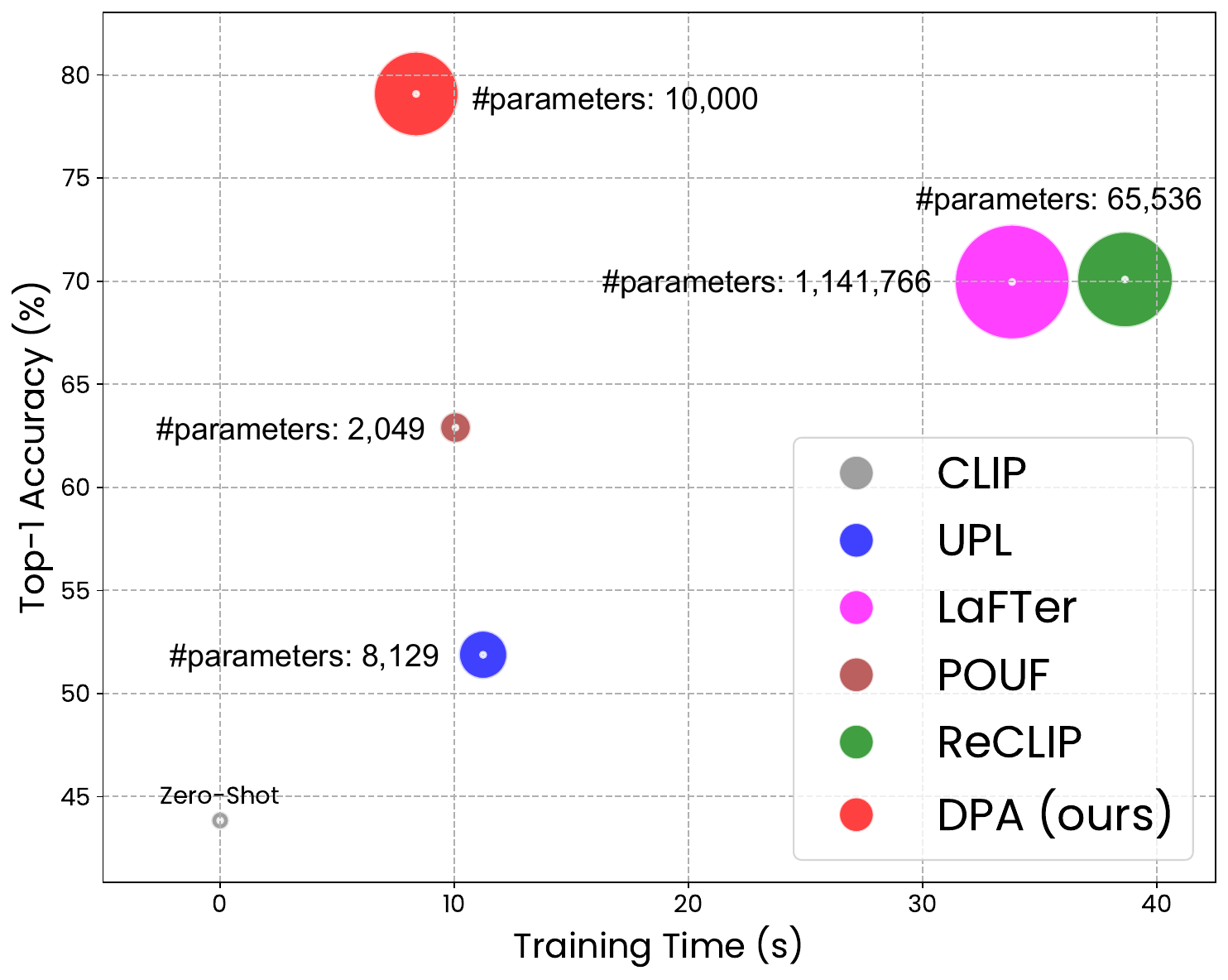}
  \caption{Efficiency comparison of \ours with baselines. The radius of each circle represents trainable parameters in each method.}
  \label{fig:run}
\end{figure}

\noindent\textbf{Comparison with ReCLIP:} We provide a comparison with a very recent method for unsupervised adaptation of VLMs, namely ReCLIP~\cite{reclip} (\Cref{table:reclip}). 
\ours consistently outperforms ReCLIP~\cite{reclip} in both inductive and transductive settings. Also, \ours has superior training efficiency and performance with fewer parameters and reduced computational complexity comparing the baselines (\Cref{fig:run}).

\begin{table}[!ht]
  \centering
  \scriptsize
  \resizebox{\columnwidth}{!}{
    \begin{tabular}{@{}lccccccccc@{}}
      \toprule
      \textbf{Method} & \textbf{Caltech} & \textbf{DTD} & \textbf{ESAT} & \textbf{FGVCA} & \textbf{Flower} & \textbf{OxPets} & \textbf{StCars} & \textbf{UCF} & \textbf{Avg} \\
      \midrule
      \textbf{Zero-shot CLIP} & 90.69 & 44.42 & 43.84 & 19.50 & 66.46 & 87.50 & 58.74 & 64.20 & 59.42 \\
      \midrule
      \rowcolor[gray]{0.8} \multicolumn{10}{c}{Inductive} \\
      \midrule
      \textbf{ReCLIP}\textsuperscript{\dag}~\cite{reclip} & 95.94 & 53.88 & 70.80 & 18.87 & 72.63 & 87.49 & 59.22 & 67.01 & 65.73 \\
      \textbf{\ours} & \textbf{96.06} & \textbf{55.69} & \textbf{80.04} & \textbf{20.67} & \textbf{75.56} & \textbf{90.71} & \textbf{62.62} & \textbf{68.49} & \textbf{68.73} \\
      \midrule
      \rowcolor[gray]{0.8} \multicolumn{10}{c}{Transductive} \\
      \midrule
      \textbf{ReCLIP}~\cite{reclip} & \textbf{92.43} & 52.50 & 59.30 & 20.34 & 70.65 & 88.42 & 59.06 & \textbf{69.13} & 63.98 \\
      \textbf{\ours} & 91.62 & \textbf{57.13} & \textbf{71.84} & \textbf{20.76} & \textbf{74.14} & \textbf{91.11} & \textbf{62.58} & 68.44 & \textbf{67.20} \\
      \bottomrule
    \end{tabular}
  }
  \caption{Comparison of top-1 classification accuracy (\%) of \ours with ReCLIP~\cite{reclip}, a source-free unsupervised domain adaptation method for CLIP. \textsuperscript{\dag} indicates that the model was originally trained in a transductive setting, but we trained it in an inductive setting for fair comparison.}
  \label{table:reclip}
\end{table}

\begin{table}[]
  \centering
  \small
  \scalebox{0.5}{
  \begin{tabular}{@{}lcccccccccccc@{}}
    \toprule
    \textbf{Meth.} & \textbf{Caltech} & \textbf{DTD} & \textbf{ESAT} & \textbf{FGVCA} & \textbf{Food} & \textbf{Flower} & \textbf{OxPets} & \textbf{StCars} & \textbf{CIF.10} & \textbf{CIF.100} & \textbf{UCF} & \textbf{Avg} \\
    \midrule
    \textbf{ZS} & 90.69 & 44.42 & 43.84 & 19.50 & 82.40 & 66.46 & 87.50 & 58.74 & 89.80 & 65.10 & 64.20 & 64.79 \\
    \midrule
    \textbf{\ours} & \textbf{96.06} & 55.69 & \textbf{80.04} & 20.67 & \textbf{84.76} & 75.56 & \textbf{90.71} & 62.62 & \textbf{95.97} & \textbf{76.47} & 68.49 & 73.37 \\
    \midrule
    \rowcolor[gray]{0.8} \multicolumn{13}{c}{\textbf{8-shot}} \\
    \midrule
    \textbf{CoOp} & 93.96 & \underline{61.82} & \underline{77.17} & \underline{26.73} & 78.84 & \underline{88.71} & 85.61 & \underline{66.20} & \underline{94.63} & \underline{75.91} & \underline{77.66} & \underline{75.20} \\
    \textbf{MaPLe} & 93.23 & 34.99 & 65.25 & 21.96 & 80.51 & 68.78 & 81.85 & 58.51 & 88.23 & 71.14 & 69.18 & 66.69 \\
    \midrule
    \rowcolor[gray]{0.8} \multicolumn{13}{c}{\textbf{16-shot}} \\
    \midrule
    \textbf{CoOp} & \underline{95.09} & \textbf{67.55} & 76.78 & \textbf{30.93} & 79.12 & \textbf{93.34} & \underline{86.70} & \textbf{70.73} & 94.50 & 75.78 & \textbf{79.51} & \textbf{77.28} \\
    \textbf{MaPLe} & 93.35 & 53.43 & 71.49 & 21.48 & \underline{81.01} & 72.47 & 86.54 & 60.28 & 91.94 & 71.74 & 70.26 & 70.36 \\
    \bottomrule
  \end{tabular}}
  \caption{Comparison with few-shot methods across 11 different datasets. \ours deliver competitive performance or outperform few-shot methods in 6 of 11 datasets.}
  \label{table:few_shot}
\end{table}

\noindent\textbf{Comparison with Few-Shot methods:} \Cref{table:few_shot} provides a performance comparison of \ours with the few-shot methods (CoOp~\cite{zhou2022learning} and MaPLe~\cite{khattakMaPLe}). \ours show competitive performance or surpasses few-shot methods in 6 of 11 datasets.

\section{Conclusion}
We present \ours, an unsupervised domain adaptation method for VLMs that aims at bridging the domain gap between visual and textual representations. \ours introduces a novel idea of dual prototypes, which work as two distinct classifiers, and their outputs are fused via a convex combination. It also ranks pseudo-labels for robust self-training during early training. \ours also enhances the alignment of visual-textual prototypes to improve adaptation performance further. \ours shows improved pseudo-labeling accuracy, which leads to notable performance improvements on various downstream tasks. \ours significantly outperforms zero-shot CLIP and the state-of-the-art methods across 13 downstream vision tasks.

{\small
\bibliographystyle{ieee_fullname}
\bibliography{egbib}
}

\end{document}


\title{\ours: Dual Prototypes Alignment for Unsupervised Adaptation of Vision-Language Models (Supplementary Material)}

\maketitle

\paragraph{Implementation and Computation Details:}
We implement our method, \ours, using the PyTorch deep learning framework~\cite{paszke2017automatic}. We use a single NVIDIA Quadro RTX6000 GPU with 24 GB  GPU memory to conduct all experiments for \ours and the baseline methods. We utilize a set of prompts from~\cite{li2022masked}, offering a variety of prompts tailored to different image classification datasets to generate textual prototypes for each dataset. 
To ensure a comprehensive evaluation, we compare our method against several state-of-the-art baselines: UPL~\cite{upl}, POUF~\cite{pouf}, LaFTer~\cite{lafter}, and ReCLIP~\cite{reclip}. For a fair comparison, we use the official codebases released by the authors, which are obtained from their respective websites. We reproduced the baseline results as the original studies did not provide outcomes using the same backbone or employed different datasets or data splits compared to those in our work. For example, LaFTer did not report results for datasets such as Stanford Cars, FGVCA, Food, or Oxford Pets. Our results are generally consistent with theirs, though they are slightly different, likely due to variations in hardware configuration. For POUF, we could not locate results for the same datasets/ or the backbones, necessitating the reproduction of the results using our datasets to ensure a fair comparison. In subsequent research, such as ReCLIP, the results for POUF were presented from a transductive training approach rather than an inductive one. UPL reported results obtained using a few-shot approach, specifically with 16 shots. This method generated pseudo-labels for the entire dataset, and then the top 16 most confident labels were selected for training. However, they only provided results for training the model using part of the dataset.
To maintain consistency across the baselines, we adapt ReCLIP to use the ViT-B-32 backbone, which is the same as the one used by \ours. This ensures that any performance differences are not attributed to variations in the model architecture. We used the same hyperparameters specified in the original papers and GitHub repositories for the baseline models, which were used to achieve the best performance for their methods, according to their papers. However, we observe that the hyperparameters reported by the baseline repositories differ from each other. UPL, POUF, LaFTer, and ReCLIP each optimize different parameters within the CLIP model, which requires them to use different hyperparameter spaces. We followed a similar approach and did not adhere to the same parameter space, as it aligns more closely with the studies in the related literature.

\paragraph{Dataset Statistics and Splits:}
\Cref{tabel:detail} provides detailed information about each dataset used in our experiments, including the number of prompts, the number of classes, the size of the training set, and the size of the testing set. This table serves as a quick reference for understanding the scale and complexity of the datasets involved in our study. Due to the massive size of ImageNet, we use only $25\%$ of the dataset (approximately 50,000 samples) to ensure efficient running times and accommodate the limitations of our computational resources. To maintain consistency, we also use the same subset of ImageNet to train the baseline methods.
Despite using a smaller portion of ImageNet, our method, \ours, still achieves comparable performance to the baselines. This demonstrates the effectiveness and efficiency of our approach, even when working with limited computational resources.

\begin{table}[!ht]
  \centering
  \small
  \begin{tabular}{@{}l@{\hskip 4pt}c@{\hskip 4pt}c@{\hskip 4pt}c@{\hskip 4pt}c@{\hskip 4pt}c@{}}
    \toprule
    \textbf{Dataset} & \textbf{Abbr.} & \textbf{Prompts} & \textbf{Classes} & \textbf{Train} & \textbf{Test} \\
    \midrule
    Caltech101 & Caltech & 34 & 100 & 4,403 & 6,645 \\
    DTD & DTD & 6 & 47 & 3,760 & 1,880 \\
    EuroSAT & ESAT & 3 & 10 & 10,000 & 5,000 \\
    FGVC Aircraft & FGVCA & 2 & 100 & 3,334 & 3,333 \\
    Food101 & Food & 1 & 101 & 75,750 & 25,250 \\
    Flowers102 & Flower & 1 & 102 & 4,093 & 2,463 \\
    Oxford Pets & OxPets & 1 & 37 & 3,680 & 3,669 \\
    SUN397 & SUN & 2 & 397 & 76,129 & 21,758 \\
    Stanford Cars & StCars & 8 & 196 & 8,144 & 8,041 \\
    CIFAR10 & CIFAR10 & 18 & 10 & 50,000 & 10,000 \\
    CIFAR100 & CIFAR100 & 18 & 100 & 50,000 & 10,000 \\
    UCF101 & UCF & 48 & 101 & 9,537 & 3,783 \\
    ImageNet & ImgNet & 130 & 1,000 & 50,000 & 50,000 \\
    ImageNet-Sketch & ImgNet-S & 7 & 1,000 & - & 50,889 \\
    ImageNet-Adversarial & ImgNet-A & 7 & 200 & - & 7,500 \\
    ImageNet-Rendition & ImgNet-R & 7 & 200 & - & 30,000 \\
    \bottomrule
  \end{tabular}
  \caption{Detailed dataset statistics.}
  \label{tabel:detail}
\end{table}

\paragraph{Comparison of Training Time and Model Complexity:}
In \cref{table:time}, we provide a comprehensive comparison between our method and the baseline approaches, focusing on two key aspects: training time and the number of parameters involved. This comparison offers a fair and balanced assessment, allowing for a clear understanding of each technique's computational efficiency and resource requirements.

\begin{table}[!ht]
  \centering
  \resizebox{\columnwidth}{!}{
    \begin{tabular}{@{}lccc@{}}
      \toprule
      \textbf{Method} & \textbf{Training Time (minutes)} & \textbf{\# Parameters} & \textbf{Top-1 Accuracy (\%)} \\
      \midrule
      Zero-shot CLIP    & 0.00  & 0         & 43.84 \\  
      UPL               & 11.23 & 8,192     & 51.88 \\  
      POUF              & 16.74 & 2,049     & 62.90 \\
      LaFTer            & 79.53 & 1,141,766 & 69.96 \\   
      ReCLIP            & 66.65 & 65,536    & 70.80 \\
      \textbf{\ours}    & 16.12 & 10,000    & 80.04 \\
      \bottomrule
    \end{tabular}
  }
  \caption{Comparison of SOTA unsupervised adaptation methods: training time, parameter count, and top-1 accuracy evaluated on the EuroSAT dataset using CLIP-ViT-B/32 as the backbone.}
  \label{table:time}
\end{table}

\paragraph{Hyperparameters Analysis:}
We outline the hyperparameters for \ours in \cref{table:hype}.

\begin{table*}[!ht]
  \centering
  \resizebox{\textwidth}{!}{%
    \begin{tabular}{@{}lccccccccccccc@{}}
      \toprule
      \textbf{Params} & \textbf{ImgNet} & \textbf{Caltech} & \textbf{DTD} & \textbf{ESAT} & \textbf{FGVCA} & \textbf{Food} & \textbf{Flower} & \textbf{OxPets} & \textbf{SUN} & \textbf{StCars} & \textbf{CIFAR10} & \textbf{CIFAR100} & \textbf{UCF} \\
      \midrule
      Batch Size & 64 & 64 & 64 & 64 & 64 & 64 & 64 & 64 & 64 & 64 & 64 & 64 & 64 \\
      Learning Rate (lr) & 1e-6 & 1e-4 & 5e-5 & 3e-5 & 3e-6 & 1e-6 & 1e-4 & 7e-5 & 1e-6 & 5e-6 & 8e-5 & 1e-5 & 1e-5 \\
      Epochs & 15 & 15 & 15 & 10 & 10 & 15 & 15 & 15 & 15 & 15 & 15 & 15 & 15 \\
      \(\beta\) & 0.5 & 0.5 & 0.2 & 0.5 & 0.5 & 0.2 & 0.5 & 0.01 & 0.01 & 0.5 & 0.1 & 0.2 & 0.5 \\
      \(\lambda_1\) & 1.0 & 1.0 & 1.0 & 1.0 & 1.0 & 1.0 & 1.0 & 1.0 & 1.0 & 1.0 & 1.0 & 1.0 & 1.0 \\
      \(\lambda_2\) & 1.0 & 1.0 & 1.0 & 1.0 & 1.0 & 1.0 & 1.0 & 1.0 & 0.2 & 1.0 & 1.0 & 1.0 & 1.0 \\
      \(\lambda_3\) & 1.0 & 1.0 & 1.0 & 1.0 & 1.0 & 1.0 & 1.0 & 1.0 & 0.01 & 1.0 & 0.1 & 1.0 & 1.0 \\
      \bottomrule
    \end{tabular}%
  }
  \caption{Hyperparameters for \ours on the downstream datasets.}
  \label{table:hype}
\end{table*}

\paragraph{With other VLMs:}
We evaluate the efficacy of \ours with other VLM models in \cref{table:diff_archit}. 
To assess this, we conducted experiments on SLIP~\cite{mu2022slip}, a version of CLIP incorporating self-supervised learning objectives during pre-training. As depicted in \cref{table:diff_archit}, \ours consistently demonstrates substantial and consistent improvements across diverse VLMs.

\begin{table*}[!ht]
  \centering
  \resizebox{\textwidth}{!}{%
    \begin{tabular}{@{}lccccccccccc@{}}
      \toprule
      \textbf{Method} & \textbf{Caltech} & \textbf{DTD} & \textbf{ESAT} & \textbf{FGVCA} & \textbf{Food} & \textbf{Flower} & \textbf{OxPets} & \textbf{StCars} & \textbf{CIFAR10} & \textbf{CIFAR100} & \textbf{UCF} \\
      \midrule
      & Init $\rightarrow$ Adapt & Init $\rightarrow$ Adapt & Init $\rightarrow$ Adapt & Init $\rightarrow$ Adapt & Init $\rightarrow$ Adapt & Init $\rightarrow$ Adapt & Init $\rightarrow$ Adapt & Init $\rightarrow$ Adapt & Init $\rightarrow$ Adapt & Init $\rightarrow$ Adapt & Init $\rightarrow$ Adapt \\ \midrule
      \textbf{SLIP} (ViT-L/16) & 83.90 $\rightarrow$ \textbf{90.20} & 25.53$\rightarrow$ \textbf{36.28} & 20.90$\rightarrow$ \textbf{31.50} & 8.40$\rightarrow$ \textbf{9.18} & 60.45$\rightarrow$ \textbf{68.08} & 64.23$\rightarrow$ \textbf{75.07} & 33.22$\rightarrow$ \textbf{44.02} & 8.00$\rightarrow$ \textbf{9.03} & 81.67$\rightarrow$ \textbf{88.10} & 48.70$\rightarrow$ \textbf{59.88} & 38.09$\rightarrow$ \textbf{52.76} \\
      \bottomrule
    \end{tabular}%
  }
  \caption{effectiveness of \ours on different model architectures and pre-training strategies.}
  \label{table:diff_archit}
\end{table*}

\begin{figure}[!ht]
  \centering
  \includegraphics[width=0.8\linewidth]{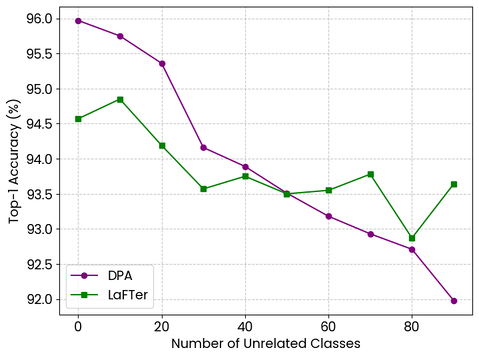}
  \caption{Top-1 accuracy ($\%$) for the CIFAR-10 dataset (using the ViT-B/32 backbone) with the inclusion of unlabeled samples from unrelated classes in the CIFAR-100 dataset. The target classes are restricted to those in the CIFAR-10 dataset only.}
  \label{fig:cifar10_noise}
\end{figure}

\paragraph{Robustness to Noisy Data:}
We conducted an experiment where we mixed unlabeled images from unrelated classes of CIFAR-100 with CIFAR-10 to add noise. Our \ours model consistently outperformed LaFTer~\cite{lafter} when adding samples of up to 50 unrelated classes, demonstrating its robustness even without using LLM descriptions (see \cref{fig:cifar10_noise}). Beyond 50 classes, LaFTer performed better, likely due to their LLM descriptions aiding task-specific performance. While our research scope doesn’t extend to domain generalization, this experiment indicates that \ours can handle noisy data effectively, showing its potential for robustness across multiple tasks.

\paragraph{Comparison with Few-shot Tip-Adapter:}
Our method, \ours, exhibits four key differences compared to Tip-Adapter~\cite{zhang2022tip}. Firstly, while the concept of image prototypes was not a primary focus in the Tip-Adapter, which utilizes a cache model of keys and values constructed from few-shot labeled samples, our approach constructs prototypes by averaging all image features across the dataset. Interestingly, if Tip-Adapter's cache size were set to 1—by averaging all samples within each class—the cache model would effectively function as a prototype. However, using a cache size of 1 notably degrades performance, as larger cache sizes yield better results~\cite{zhang2022tip}. Secondly, the cache model in Tip-Adapter is constructed based on true labels, as it primarily focuses on the few-shot labeled adaptation of CLIP, whereas our method utilizes pseudo labels to construct the prototypes, which are updated in an online manner throughout the training process. The method of updating prototypes also differs significantly, as in Tip-Adapter, the cache model remains fixed and is directly used for inference on the test dataset, with even the optimized version, Tip-Adapter-F~\cite{zhang2022tip}, only optimizing the keys of the cache model using stochastic gradient descent (SGD). In contrast, our approach updates the image prototypes at the end of each epoch by leveraging the image features and the pseudo labels stored in the memory bank. Finally, \ours employs two types of prototypes—image and textual prototypes—to generate pseudo labels, which is not the case in Tip-Adapter, and incorporates three critical components: center, weighting (w), and alignment, which are absent in Tip-Adapter and significantly enhance adaptation performance.

\paragraph{Detailed Analysis of Model Components}
\noindent While the improvements from weighting and alignment are not as significant as those from the center component, as shown in Table 3 of the main paper, it is important to note that both weighting and alignment enhance performance across various datasets, including Caltech101, EuroSAT, FGVCA, Stanford Cars, and UCF. Our findings indicate that the weighting strategy significantly improves performance on specific datasets, such as EuroSAT (+9.50\%). We assert that the EuroSAT dataset presents complex and abstract tasks, resulting in more misalignment than other datasets. This complexity causes noisy pseudo-labels generated from image prototypes, resulting in lower performance at the beginning of training, as demonstrated in Figure 4 of the main paper. This weighting approach adjusts the influence of pseudo-labels based on their confidence levels, reducing the impact of noise in the early stages of training. Furthermore, when weighting is combined with alignment, we observe substantial improvements across several datasets, including Caltech101(+0.62\%), FGVCA(+0.87\%), Stanford Cars(+1.09\%), and UCF(+1.19\%).

\paragraph{Algorithms:}
As outlined in the Methodology Section of the main paper, \ours fine-tunes solely the layer-normalization weights in CLIP's visual encoder $E_v$ and the prompt-initialized textual prototypes $\textbf{Z}$. In this section, we present the pseudocode for the initialization of textual prototypes in \cref{alg:unclif_text_init}, the self-training process utilizing dual prototype PL generation in \cref{alg:unclif_train}, and the inference procedure for \ours in \cref{alg:inference}.

\begin{algorithm}[!ht]
    \caption{Initializing the textual prototypes}\label{alg:unclif_text_init}
    \begin{algorithmic}[1]
        \Require Text encoder $E_t$, Target class names $\mathcal{C}_t = \{c^j\}_{j=1}^C$, Prompts $\{\mathbf{x}_{prompt}^i\}_{i=1}^k$
        \Ensure $E_t$ is pre-trained
        \Function{init-text-prototypes}{$E_t$, $\mathcal{C}_t$, Prompts}
            \State Initialize $\mathbf{Z} \in \mathbb{R}^{C \times d} \gets \mathbf{0}$
            \For{$j=1, \dots, C$}
                \State $\mathbf{Z}_j \gets \text{mean}\left(\left\{E_t\left(p_i \oplus c^j\right) \ \text{for} \ i=1, \dots, k\right\}\right)$
            \EndFor
            \State \Return $\mathbf{Z}$
        \EndFunction
    \end{algorithmic}
\end{algorithm}

\begin{algorithm*}[!ht]
    \caption{\ours self-training}\label{alg:unclif_train}
    \resizebox{\textwidth}{!}{%
    \begin{minipage}{\textwidth}
        \begin{algorithmic}[1]
            \Require Vision encoder of a source vision-language model $E_v^\Theta$ parameterized by its trainable parameters $\Theta$,
            \Statex \hspace{2.5em} Unlabelled images from the target dataset $\mathcal{X}_t = \{x_i\}_{i=1}^N$,
            \Statex \hspace{2.5em} Initialized text prototypes $\mathbf{Z}$,
            \Statex \hspace{2.5em} Weak augmentation $\alpha(\cdot)$,
            \Statex \hspace{2.5em} Strong augmentation $\mathcal{A}(\cdot)$,
            \Statex \hspace{2.5em} Number of epochs $\texttt{MaxEpochs}$,
            \Statex \hspace{2.5em} Batch size $\texttt{B}$
            \Ensure $E_v$ has pre-trained model parameters $\Theta$
            \Function{ComputeImageCentroids}{$E_v^\Theta$, $\mathbf{Z}$, $\mathcal{X}_t$}
                \State $\hat{y} \gets \left\{\argmax_j E_v\left(x_i\right) \cdot \mathbf{Z}^T\right\}_{i=1}^N \in \mathbb{Z^+}^{N\times 1}$
                \State $\mathbf{F} \gets \left\{\left\{\mathbb{I}_{\hat{y_i}=c_j}E_v^\Theta\left(x^i\right)\right\}_{i=1}^N\right\}_{j=1}^C$ \Comment{Shown separately for the convenience of following notation}
                \State $\mathbf{P}_v \gets \left\{\frac{1}{|\mathbf{F}_j|}\sum_{i=1}^N\mathbf{F}_{ji}\right\}_{j=1}^C$
                \State \Return $\mathbf{P}_v$
            \EndFunction
            \State $\mathbf{P}_v \gets$ \Call{ComputeImageCentroids}{$E_v^\Theta$, $\mathbf{Z}$, $\mathcal{X}_t$} \Comment{Initial image prototypes}
            \State $\textbf{MB} \gets \{\}$ \Comment{Initialize the memory bank}
            \For{$\texttt{epoch} \gets 1$ to $\texttt{MaxEpochs}$}
                \State $\textbf{x} \gets$ \Call{SampleMiniBatch}{$\mathcal{X}_t$, $B$} \Comment{$\mathbf{x} \in \mathbb{R}^{B\times d}$}
                \State With no Back-Propagation:
                \State \hspace{2.5em} $\mathbf{f} \gets E_v^\Theta\left(\alpha(\textbf{x})\right)$ \Comment{Weakly augmented mini-batch of image features $\in \mathbb{R}^{B\times d}$}
                \State \hspace{2.5em} $\mathbf{p}_t \gets softmax\left(\mathbf{f} \cdot \mathbf{Z}^T, axis=1\right)$ \Comment{Textual pseudo-probabilities $\in \mathbb{R}^{B\times C}$}
                \State \hspace{2.5em} $\mathbf{p}_v \gets softmax\left(\mathbf{f} \cdot \mathbf{P}_v^T, axis=1\right)$ \Comment{Visual pseudo-probabilities $\in \mathbb{R}^{B\times C}$}
                \State \hspace{2.5em} $\hat{\mathbf{p}} \gets \beta \cdot DA(\mathbf{p}_t) + (1-\beta) \cdot \mathbf{p}_v$ \Comment{Final pseudo-probabilities}
                \State \hspace{2.5em} $\mathbf{\hat{y}} \gets \argmax_j \hat{\mathbf{p}}$ \Comment{Final PLs for the mini-batch}
                \State \hspace{2.5em} $\mathbf{w_x} \gets \langle \mathbf{f}, \mathbf{Z}(\mathbf{\hat{y}}) \rangle \langle \mathbf{f}, \mathbf{Z}(\mathbf{\hat{y}}) \rangle$
                \State \hspace{2.5em} Update $\mathbf{MB}$ with $\{\mathbf{f}_b, \mathbf{\hat{y}_b}\}_{b=1}^B$
                \State \hspace{2.5em} $\mathbf{P}_v \gets$ \Call{ComputeImageCentroids}{$E_v^\Theta$, $\mathbf{Z}$, $\mathcal{X}_t$} \Comment{Update image prototypes}
                \State $p_{\mathcal{A}(x)} \gets softmax(E_v^\Theta\left(\mathcal{A}(\mathbf{x})\right) \cdot \mathbf{Z}^T, axis=1)$ \Comment{Strongly-augmented counterpart}
                \State $\mathcal{L}_{st} \gets \mathbf{w_x} \cdot cross\_entropy(p_{\mathcal{A}(x)}, \mathbf{\hat{y}})$ \Comment{Self-training loss}
                \State $\mathcal{L}_{reg} \gets -\frac{1}{C} \sum_{j=1}^C log\left(\bar{p}_{\mathcal{A}(x),j}\right)$ \Comment{Fairness regularization loss}
                \State $\mathcal{L}_{align} \gets cross\_entropy(\mathbf{P} \cdot \mathbf{Z}^T / \tau, arange(C))$ \Comment{Prototypes alignment loss}
                \State $\mathcal{L} \gets \lambda_1\mathcal{L}_{st} + \lambda_2\mathcal{L}_{reg} + \lambda_3\mathcal{L}_{align}$
                \State Back-Propagate over $\Theta$ and $\mathbf{Z}$ on $\mathcal{L}$
            \EndFor
        \end{algorithmic}
    \end{minipage}
    }
\end{algorithm*}

\begin{algorithm}[H]
    \caption{\ours Inference}\label{alg:inference}
    \begin{algorithmic}[1]
        \Require Vision encoder $E_v$, Test image dataset $x$, Text prototypes $\mathbf{Z}$, Weak augmentation $\alpha(\cdot)$
        \Ensure $E_v$ and $\mathbf{Z}$ have fine-tuned \ours parameters
        \Function{\ours}{$E_v$, $\mathbf{Z}$, $\alpha(\cdot)$, $x$}
            \State $\hat{y} \gets \argmax_j E_v\left(\alpha(x)\right) \cdot \mathbf{Z}^T$
            \State \Return $\hat{y}$
        \EndFunction
    \end{algorithmic}
\end{algorithm}

\noindent\textbf{Limitation and Future Work:}
As presented in Table 6 of the main paper, we observe significant performance improvements under distribution shift in 2 out of 3 datasets. This enhancement is tied to the challenges that CLIP encounters when fine-tuned on ImageNet and subsequently evaluated on out-of-distribution datasets, where domain shifts are introduced. Fine-tuning on ImageNet can undermine CLIP robustness, as highlighted in previous studies~\cite{clip, li2022masked}. In contrast to CoOp~\cite{zhou2022learning}, a few-shot supervised adaptation method, our approach consistently improves performance across all three out-of-distribution datasets. These results align with findings from CLIP adaptation literature and underscore the robustness of our adaptation strategy~\cite{clip, li2022masked}. Future work will focus on further enhancing this robustness.

\begin{figure*}[!ht]
    \centering
    \begin{subfigure}[b]{0.32\textwidth} 
        \centering
        \includegraphics[width=\textwidth]{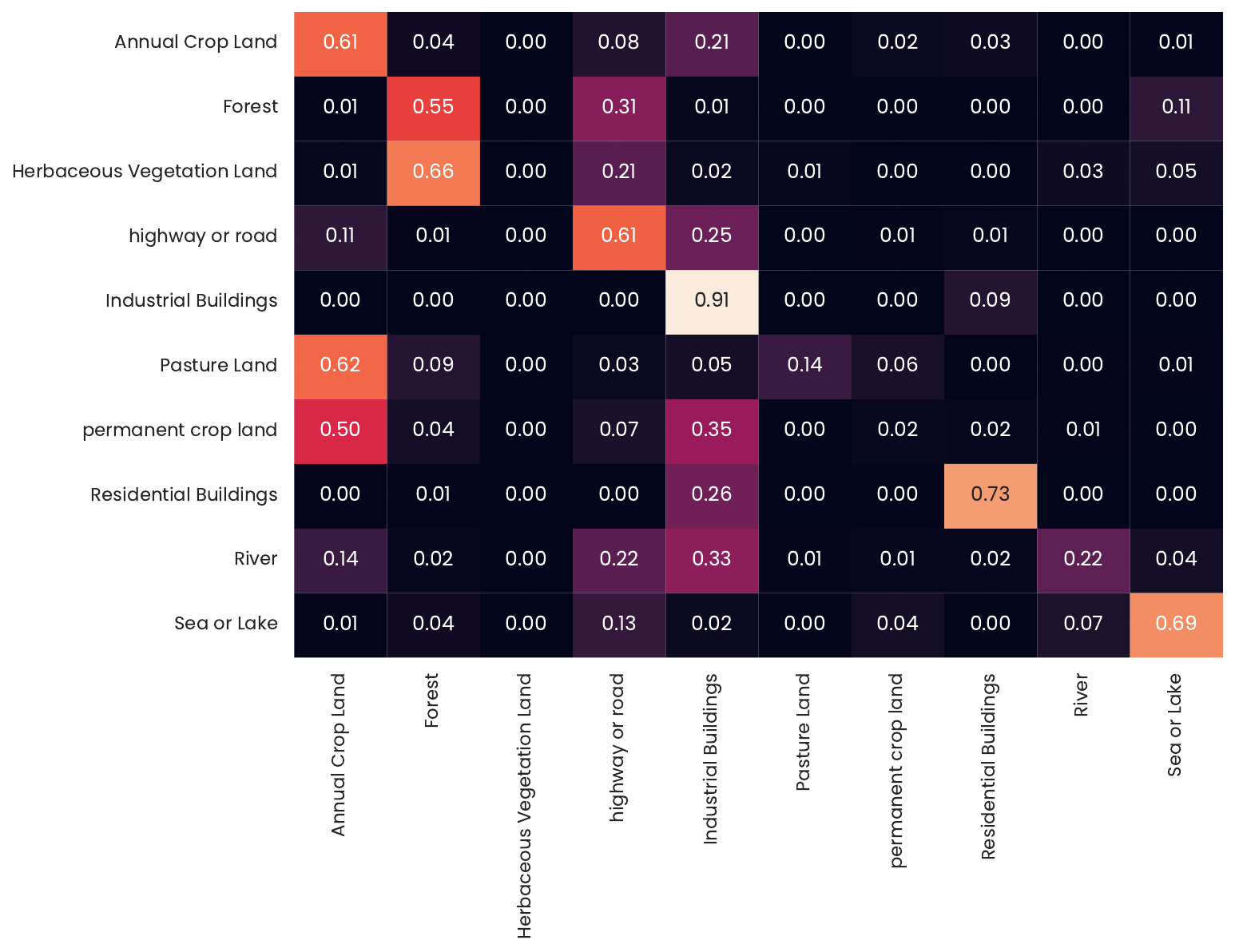}
        \caption{Zero-shot CLIP}
        \label{fig:conf_mat_zs_CLIP}
    \end{subfigure}
    \hfill
    \begin{subfigure}[b]{0.32\textwidth} 
        \centering
        \includegraphics[width=\textwidth]{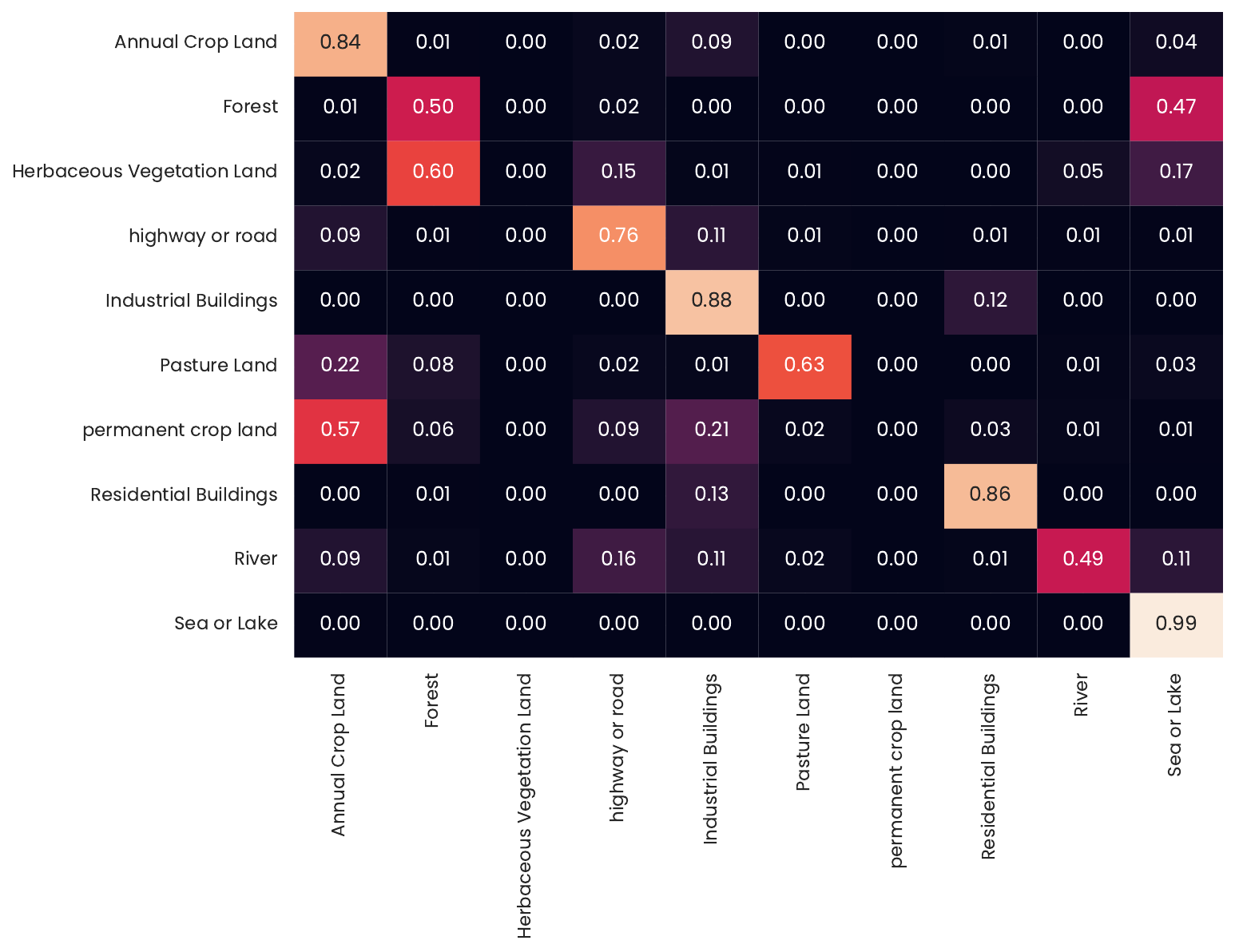}
        \caption{ReCLIP (inductive training)}
        \label{fig:conf_mat_ReCLIP}
    \end{subfigure}
    \hfill
    \begin{subfigure}[b]{0.32\textwidth} 
        \centering
        \includegraphics[width=\textwidth]{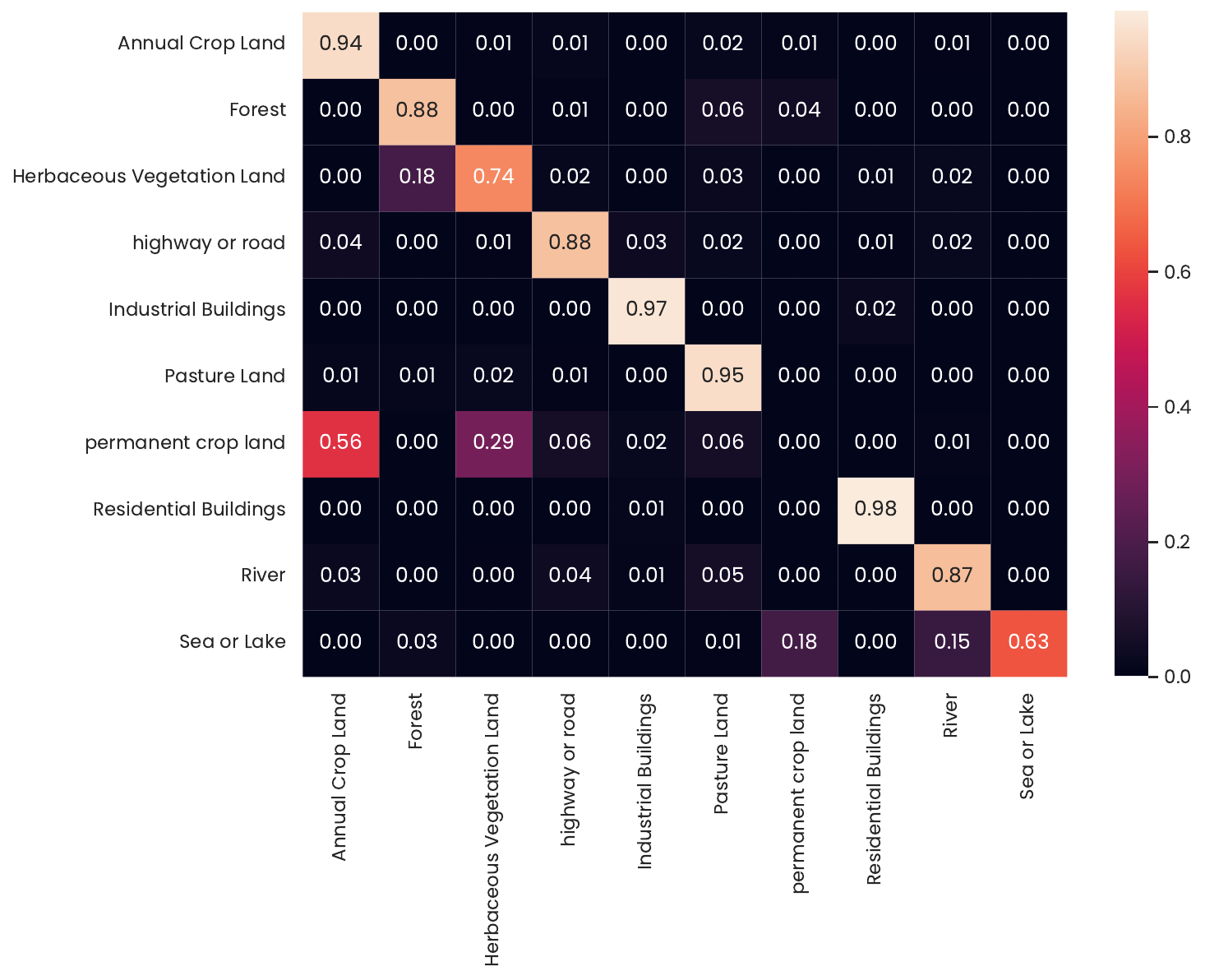}
        \caption{\ours (Ours)}
        \label{fig:conf_mat_ours}
    \end{subfigure}
    \caption{Confusion matrices of zero-shot CLIP, ReCLIP~\cite{reclip}, and our method (\ours).}
    \label{fig:conf_mats}
\end{figure*}

Additionally, while our method shows promising results, there remains room for improvement in mitigating zero-shot CLIP's strong confirmation biases, as illustrated in \cref{fig:conf_mats}, which was not the primary focus of this work. This figure highlights the remarkable effectiveness of our approach in source-free adaptation using dual classifiers to generate pseudo-labels to reduce false positive rates across all categories. However, zero-shot CLIP has strong confirmation biases, resulting in our method consistently achieving false positive predictions for certain classes. The confusion between classes is also evident in ReCLIP~\cite{reclip}, indicating that addressing these strong biases requires more robust de-biasing strategies.

{\small
\bibliographystyle{ieee_fullname}
\bibliography{egbib}
}